\documentclass{article} 
\usepackage{iclr2027_conference,times}

\usepackage{amsmath,amsfonts,bm}

\def\eqref#1{equation~\ref{#1}}

\def\1{\bm{1}}
\newcommand{\train}{\mathcal{D}}

\def\vc{{\bm{c}}}

\def\vf{{\bm{f}}}

\def\vo{{\bm{o}}}
\def\vp{{\bm{p}}}
\def\vq{{\bm{q}}}

\def\vs{{\bm{s}}}
\def\vt{{\bm{t}}}
\def\vu{{\bm{u}}}

\def\vx{{\bm{x}}}

\def\mD{{\bm{D}}}

\def\mK{{\bm{K}}}

\def\mR{{\bm{R}}}

\def\mW{{\bm{W}}}

\def\mSigma{{\bm{\Sigma}}}

\DeclareMathAlphabet{\mathsfit}{\encodingdefault}{\sfdefault}{m}{sl}
\SetMathAlphabet{\mathsfit}{bold}{\encodingdefault}{\sfdefault}{bx}{n}

\newcommand{\Ls}{\mathcal{L}}
\newcommand{\R}{\mathbb{R}}

\usepackage{hyperref}
\usepackage{etoc} 
\usepackage{url}
\usepackage{graphicx}
\usepackage{wrapfig}
\usepackage{booktabs}
\usepackage{multirow}
\usepackage{amsmath}
\usepackage{amsfonts}
\usepackage{amssymb}
\usepackage{amsthm}
\usepackage{array}
\usepackage{makecell}
\usepackage{xspace}
\usepackage[capitalize]{cleveref}

\newcommand{\ours}{LoCoSplat\xspace}
\newcommand{\volsplat}{VolSplat\xspace}
\newcommand{\tabfont}{\scriptsize\setlength{\tabcolsep}{3pt}} 
\theoremstyle{remark}
\newtheorem{remark}{Remark}
\usepackage{xcolor}

\title{LoCoSplat: Real-Time Feed-Forward\\3D Gaussian Splatting\\with Minimal 3D Reasoning}

\author{Sinan Wang$^{1*}$, Jinjin He$^{1*}$, Yuchen Sun$^{1}$, Duowen Chen$^{1}$, Shenyifan Lu$^{1}$, Bo Zhu$^{1}$ \\
\mdseries $^{1}$Georgia Institute of Technology \\
\mdseries\texttt{swang3081@gatech.edu}
}

\iclrfinalcopy 
\begin{document}

\maketitle
{\renewcommand{\thefootnote}{\fnsymbol{footnote}}\footnotetext[1]{Equal contribution.}}
\suppressfloats[t] 

\etocdepthtag.toc{main} 
\begin{abstract}
Feed-forward 3D Gaussian Splatting (3DGS) increasingly aggregates multi-view evidence with heavy learned 3D networks. We propose \ours (\textbf{Lo}cal-\textbf{Co}ntext \textbf{Splat}ting), motivated by the observation that a Gaussian is a local primitive: once depth is predicted, what the 3D stage must add (scale, rotation, opacity) depends on the point cloud around each anchor, and a fixed local average of that neighbourhood is enough to supply it, no heavy network required. \ours realises exactly this average: it splats a 16-d linear projection of the point features into a fine and a coarse grid and reads both back at each anchor with a 0.14M-parameter pointwise MLP; with no learned 3D network and no dynamic sparse computation, its whole encoder runs as one fp16 CUDA graph. On RealEstate10K, \ours outperforms every prior feed-forward method on PSNR, SSIM, and LPIPS at 6, 12, and 24 views, with a margin that widens as views densify ($+3.3$ PSNR over \volsplat, the prior voxel-aligned state of the art, at 24 views) and grows further under zero-shot transfer to ACID and fine-tuning on ScanNet. It reconstructs a 6-view scene in 33\,ms on one NVIDIA RTX PRO 6000 GPU, the fastest of seven feed-forward methods and $4.2\times$ faster than the previous state of the art, trains $2.7\times$ faster ($5.9\times$ at 24 views), and uses $6.7\times$ less inference memory.
\end{abstract}

\section{Introduction}
\label{sec:introduction}

Feed-forward 3D Gaussian Splatting~\citep{charatan2024pixelsplat,chen2024mvsplat,xu2025depthsplat} predicts the Gaussian primitives of 3DGS~\citep{kerbl20233d} from a handful of posed images in a single forward pass, replacing the per-scene optimization of NeRF and 3DGS~\citep{mildenhall2021nerf,kerbl20233d} with a learned prior. Pixel-aligned prediction, the dominant approach, suffers from multi-view misalignment and resolution-bound Gaussian budgets, and the field's answer has been to reason with ever-heavier learned modules, mostly by moving prediction into 3D: sparse 3D U-Nets and CNNs~\citep{wang2025volsplat,zhang2022nerfusion}, 3D cost-volume networks, or long-sequence Mamba/transformer models over image tokens~\citep{ziwen2024long}. \volsplat~\citep{wang2025volsplat}, the state of the art of the depth-guided voxel-aligned family, predicts depth, voxelizes the unprojected pixels into anchors, and then \emph{refines} the sparse volume with a 17M-parameter sparse 3D U-Net before decoding Gaussians. That U-Net is its largest inference cost ($\sim$7 scenes per second): a dynamic-shape sparse computation~\citep{choy20194d} that cannot be captured in a CUDA graph and barely benefits from mixed precision, and the dominant training memory as views grow (24-view training takes 159 hours). Behind this trajectory lies an implicit assumption: that aggregating multi-view evidence in 3D requires substantial learned 3D capacity.

\begin{figure}[t]
  \centering
  \includegraphics[width=\linewidth]{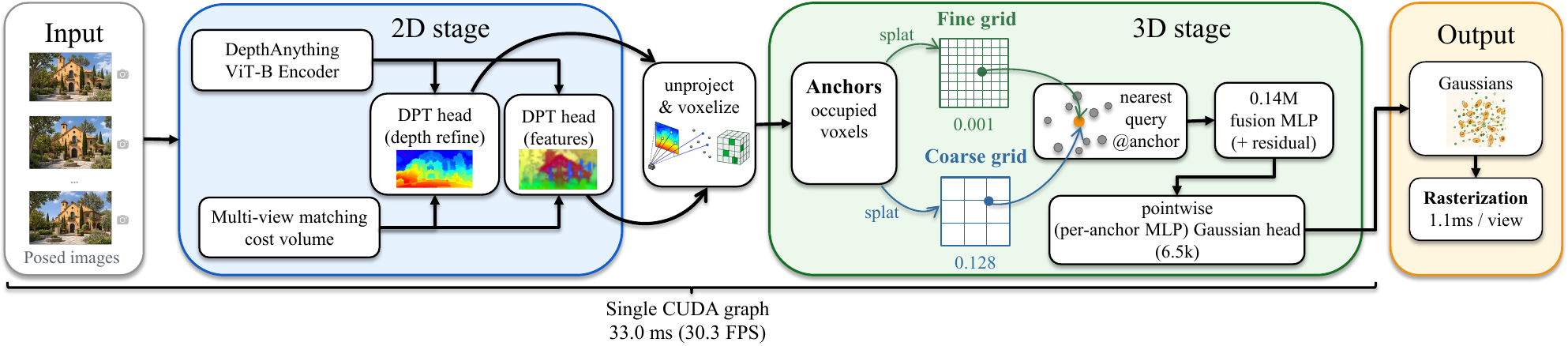}
  \caption{\textbf{\ours overview.} Predicted depth and per-pixel features are unprojected and voxelized into anchors; the 3D stage averages anchor features in a fine and a coarse grid and decodes each anchor pointwise. The whole encoder runs as one CUDA graph.}
  \label{fig:pipeline}
\end{figure}

\textbf{Our motivation is a local view of the 3D stage: what a Gaussian needs from its neighbours is the kind of low-order local information that a grid computes for free, not a learned 3D network.} \ours (\textbf{Lo}cal-\textbf{Co}ntext \textbf{Splat}ting, \cref{fig:pipeline}) keeps the depth-guided voxel-aligned pipeline (\cref{sec:prelim}) up to the anchors: a hybrid monocular--multi-view depth branch and a DPT feature head place one anchor per occupied voxel. Its 3D stage is a fixed grid average: each anchor's feature, linearly projected to 16 dimensions, is averaged into a fine and a coarse grid, and both grids are read back at the anchor by a 0.14M-parameter pointwise MLP and a pointwise head. With no learned 3D network and no dynamic sparse computation, the whole encoder runs as one fp16 CUDA graph.

The motivation is structural: \textbf{a 3D Gaussian is a local primitive}. Once depth is predicted, each anchor's colour and position are determined by its own pixel. The 3D stage must add scale, rotation, and opacity, the layout of the surface around the anchor; no single pixel observes this layout, but it is determined by the neighbouring points, which a \emph{fixed} average over a coarser neighbourhood, computed for free by a grid, can summarize. The anchors are also so sparse at the working voxel size (99.6\% have no occupied neighbour at 6 views) that a $3^3$ sparse convolution over them is already pointwise (\cref{sec:head}), so any context must come from a coarser scale, which the coarse grid supplies. This intuition led to the design.

We then measure what the 3D stage of this pipeline actually needs (\cref{sec:experiments}): very little. A fixed grid average read by a pointwise MLP of under 0.2M parameters beats the 17M-parameter U-Net by $+0.70$ PSNR (\cref{tab:evolution}). \ours has the best PSNR, SSIM, and LPIPS of all feed-forward methods on RealEstate10K at 6, 12, and 24 views, with a margin that \emph{grows} on ACID and ScanNet, and reconstructs a 6-view scene in \textbf{33.0\,ms} (\textbf{30.3 FPS}) on one GPU, $4.2\times$ faster than \volsplat ($2.4\times$ at equal fp32 precision) with 85\% less peak inference memory, and trains $2.7\times$ faster ($5.9\times$ at 24 views; \cref{tab:efficiency}).

\noindent Our contributions are:
\begin{list}{$\bullet$}{\leftmargin=0pt \itemindent=1em \labelwidth=0.7em \labelsep=0.3em \topsep=2pt \itemsep=1pt \parsep=0pt}
    \item \textbf{A 3D stage with no learned 3D network,} motivated by the locality of a Gaussian, with controlled experiments showing that the 3D stage of this pipeline needs very little: a fixed grid average read by a pointwise MLP of under 0.2M parameters beats a 17M-parameter sparse 3D U-Net, and the minimal coarse context it keeps needs neither learning nor appearance content (\cref{sec:collapse}).
    \item \textbf{\ours, a state-of-the-art real-time feed-forward 3DGS method,} leading at 6, 12, and 24 views, in and out of domain, whose fp16 CUDA-graph encoder reconstructs at 30 FPS on one GPU, $4.2\times$ faster than \volsplat with $6.7\times$ less inference memory and $2.7\times$ faster training (\cref{sec:realtime}).
    \item \textbf{A predictive account of the dense-view collapse} of depth-guided reconstruction: an optimization failure of the depth branch governed by the per-candidate plane-sweep disparity; \ours does not collapse and leads \volsplat by $+3.3$ PSNR at 24 views (\cref{sec:ablations}).
\end{list}

\section{Method}
\label{sec:method}
\label{sec:final}

\ours reconstructs a set of 3D Gaussians from $V$ posed images in one forward pass (\cref{fig:pipeline}). Its 2D stage (\cref{sec:prelim}) is a depth-guided pipeline in the \volsplat/DepthSplat family~\citep{wang2025volsplat,xu2025depthsplat}: per-view depth and per-pixel features, unprojected and voxelized into anchors (\cref{sec:anchors}). Its 3D stage (\cref{sec:theory}) has no learned 3D network: a fixed local average of the point features is read by a pointwise MLP, and a pointwise head decodes each anchor's Gaussian (\cref{sec:head}); \cref{sec:training} gives the training objective.

\begin{figure}[t]
  \centering
  \includegraphics[width=\textwidth]{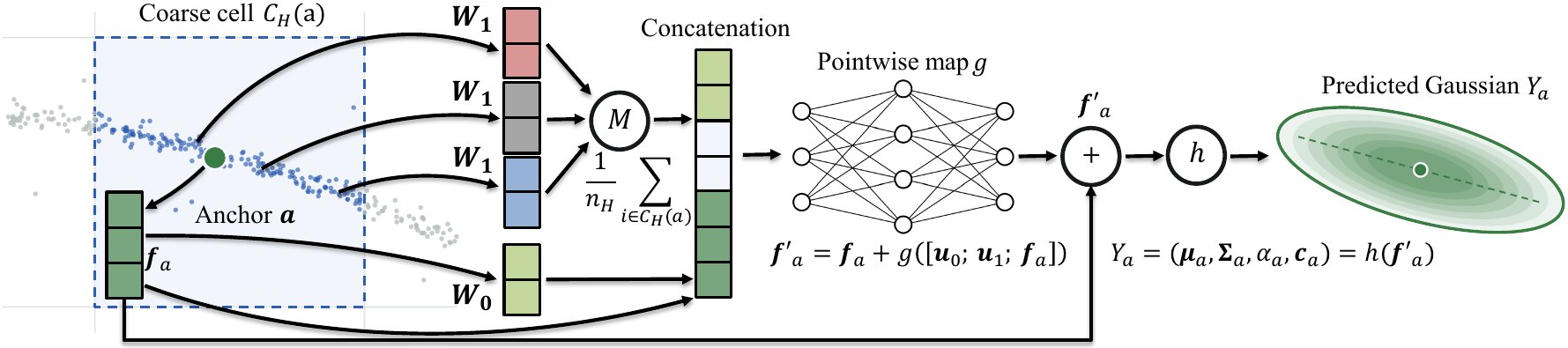}
  \caption{\textbf{Local context as a fixed average.} Points in the coarse cell $C_H(a)$ are projected by $\mW_1$ and averaged, with no parameters, into $\vu_1(a)$; the anchor's feature gives $\vu_0(a)=\mW_0\vf_a$. The pointwise map $g$ fuses $[\vu_0;\vu_1;\vf_a]$ residually and the head $h$ decodes $Y_a$ (\cref{eq:splat,eq:fuse,eq:head}).}
  \label{fig:theory}
\end{figure}

\subsection{Depth and Feature Prediction}
\label{sec:prelim}

The input is $V$ images $\mathbf I_i\in\R^{H\times W\times3}$ with intrinsics $\mK_i$ and camera-to-world poses $(\mR_i,\vt_i)$. A frozen DepthAnythingV2 ViT-B backbone~\citep{depth_anything_v2} provides monocular features, mapped by a trainable linear \emph{interface} from 768 to 384 channels (ViT-S width); a CNN with a local-window cross-view transformer over the two nearest views~\citep{xu2023unifying} provides matching features, from which a plane-sweep cost volume over $D{=}64$ inverse-depth candidates $\{d_k\}$ is built. A depthwise-separable, half-width UniMatch-style regressor turns it into matching probabilities $p_i[k,\vp]$ and the coarse inverse depth $\tilde d_i(\vp)=\sum_k p_i[k,\vp]\,d_k$; one DPT head~\citep{depth_anything_v2} refines it with the monocular features into the full-resolution depth $\mD_i$, and a second DPT head yields the per-pixel feature $\vf_{i,\vp}\in\R^{132}$ (image, depth, and peak matching probability appended) that enters the 3D stage.

\subsection{Anchors}
\label{sec:anchors}

Every pixel is unprojected along its depth and voxelized at the fine voxel size $v{=}0.001$:
\begin{equation}
\label{eq:unproject}
  \vx_{i,\vp}=\mR_i\,\mD_i(\vp)\,\mK_i^{-1}\begin{bmatrix}\vp\\1\end{bmatrix}+\vt_i,
  \qquad
  c_v(\vx)=\operatorname{rnd}(\vx/v)\in\mathbb Z^3 .
\end{equation}
This gives the point cloud $P=\{(\vx_i,\vf_i)\}_{i\le n}$, $n=V{\cdot}H{\cdot}W$, and the anchor set $\mathcal A=\{c_v(\vx_i)\}$ of occupied voxels, one Gaussian per anchor; anchor $a$ carries the mean position $\vx_a$ and mean feature $\vf_a$ of its points. At $v=0.001$ almost every occupied voxel holds a single point (\cref{tab:speed_detail}) and at 6 views only $0.42\%$ of anchors have an occupied $3^3$ neighbour, so anchors are points and the fine lattice carries no neighbourhood structure: any context the 3D stage uses must come from a coarser scale.

\subsection{Local-Context Aggregation}
\label{sec:theory}

\paragraph{The role of the 3D stage.}
Write $Y_a=(\boldsymbol\mu_a,\mSigma_a,\alpha_a,\vc_a)$ for the Gaussian of anchor $a$. Its colour and position are functions of the anchor's own pixel, $(\vf_a,\vx_a)$. Scale, rotation, and opacity describe the surface \emph{around} $a$, which no single pixel observes; they are what the 3D stage has to supply, and they depend on the layout of $P$ around $a$. Since anchors are isolated at the fine voxel size (\cref{sec:anchors}), this information has to be gathered at a coarser scale $H$, and the simplest fixed operator that does so is a local average. With $w_i(a)=K_H(\vx_a,\vx_i)$ for a kernel of bandwidth $H$ and a linear projection $\mW$ of the point features, splatting $\mW\vf_i$ into a grid of cell size $H$ and reading the cell at $a$ computes the Nadaraya--Watson estimate~\citep{nadaraya1964estimating,watson1964smooth} in its binned form,\footnote{Binning here is a computational device, the same grid accumulation used by binned kernel estimators~\citep{hall1996binned}. That analysis concerns density estimation, and we do not rely on it for a guarantee on the regression estimate in \cref{eq:nw}; what the average is worth is measured empirically in \cref{sec:collapse}.}
\begin{equation}
\label{eq:nw}
  \vu_H(a)=\frac{\sum_i w_i(a)\,\mW\vf_i}{\sum_i w_i(a)} .
\end{equation}
Because $\vf_i$ carries each point's depth and peak matching probability alongside its image features, $\mW$ can route geometric information about the neighbourhood (such as its mean depth) into $\vu_H$ as well as appearance. The motivation for \ours is that this fixed average, together with the anchor's own feature, gives a pointwise map enough context to predict the Gaussian without a learned 3D network (\cref{fig:theory}). \Cref{sec:collapse} measures what this context is worth, whether it needs to be learned or to carry appearance (by replacing $\vu_H$ with an untrained low-pass or with geometric statistics of the cell), and how well a learned 3D refiner trains in this pipeline.

\paragraph{Two grids and a pointwise map.}
\ours instantiates \cref{eq:nw} with the box kernel at two scales, with a learned 16-d projection per scale, $\mW_0\vf_i$ and $\mW_1\vf_i$. The \emph{fine} level is the base voxelization of \cref{sec:anchors} itself: its grid is the anchor lattice $c_v$, and $\vu_0(a)=\mW_0\vf_a$ is the projection of the anchor's own mean feature. The \emph{coarse} level is a grid of cell size $H=0.128$ with its own cell index $c_H(\vx)=\operatorname{rnd}(\vx/H)\in\mathbb Z^3$, the same rounding rule as $c_v$ so that cell centres sit at integer multiples of $H$; writing $n_H(c)$ for the number of points in coarse cell $c$,
\begin{align}
  \text{project \& splat:}&\qquad \vf_i\mapsto\mW_1\vf_i,\qquad\quad\ \
  \vu_1(c)=\frac{1}{n_H(c)}\sum_{i:\,c_H(\vx_i)=c}\mW_1\vf_i,\label{eq:splat}\\
  \text{query \& fuse:}&\qquad \vu_1(a)=\vu_1\big(c_H(\vx_a)\big),\qquad
  \vf'_a=\vf_a+g\big([\vu_0(a);\vu_1(a);\vf_a]\big),\label{eq:fuse}
\end{align}
with $\mW_0,\mW_1\in\R^{16\times132}$, $g:\R^{164}\to\R^{256}\to\R^{256}\to\R^{132}$, and $\vu_1(c)=\mathbf 0$ for empty cells. The splat in \cref{eq:splat} is $\vu_H$ of \cref{eq:nw} with $w_i(a)=\mathbf 1[c_H(\vx_i)=c_H(\vx_a)]$. The fine level needs no splat: a nearest-cell splat of $\mW_0\vf_i$ at cell size $v$ under the same rounding rule, read back at the anchor, returns exactly the per-voxel mean $\mW_0\vf_a$ (a linear map commutes with the mean, and each voxel's mean point lies inside its voxel), so $\vu_0$ is taken from the base voxelization directly and only the coarse level is splatted. The aggregation has no parameters and no data-dependent shape; the 3D stage's only learned parameters are $(\mW_0,\mW_1)$ (4.3k) and $g$ (0.14M).

\subsection{Gaussian Prediction}
\label{sec:head}

A two-layer pointwise MLP $h:\R^{132}\to\R^{38}\to\R^{38}$ (6.5k parameters) maps $\vf'_a$ to raw parameters $(\bar\alpha_a,\bar{\vo}_a,\bar{\vs}_a,\bar{\vq}_a,\bar{\vc}_a)\in\R^{1+3+3+4+27}$, decoded as
\begin{equation}
\label{eq:head}
\begin{aligned}
  \boldsymbol\mu_a&=\vx_a+3v\,\big(\sigma(\bar{\vo}_a)-\tfrac12\big), &
  \alpha_a&=\sigma(\bar\alpha_a),\\
  \vs_a&=\operatorname{clip}\big(\operatorname{softplus}(\bar{\vs}_a-4),\,s_{\min},\,s_{\max}\big), &
  \mSigma_a&=\mR(\bar{\vq}_a/\|\bar{\vq}_a\|)\,\operatorname{diag}(\vs_a)^2\,\mR(\bar{\vq}_a/\|\bar{\vq}_a\|)^{\!\top},
\end{aligned}
\end{equation}
where $\sigma$ is the sigmoid, $\mR(\cdot)$ the rotation matrix of a unit quaternion, and the offset is confined to $\pm1.5$ fine voxels around the anchor. The colour $\vc_a$ is a degree-2 spherical-harmonics vector whose DC term is set from the mean RGB of the anchor's pixels; $h$ predicts the higher orders. A pointwise head suffices because $99.6\%$ of anchors have no occupied $3^3$ neighbour (\cref{sec:anchors}), so a sparse convolution would reduce to its centre tap at almost every anchor (\cref{rem:pointwise}); it also keeps the pipeline free of sparse-tensor operations.

\subsection{Training Objective}
\label{sec:training}

Following DepthSplat and \volsplat~\citep{xu2025depthsplat,wang2025volsplat}, the network is trained end to end from RGB supervision alone. Writing $\hat{\mathbf I}^{(m)}_{\mathcal G}$ for the $m$-th of $M$ target views rendered from a Gaussian set $\mathcal G$ with a differentiable rasterizer~\citep{kerbl20233d},
\begin{equation}
\label{eq:loss}
  \Ls=\sum_{m=1}^{M}\Big[\ell\big(\hat{\mathbf I}^{(m)}_{\mathcal G(\mD)},\mathbf I^{(m)}\big)+\beta\,\ell\big(\hat{\mathbf I}^{(m)}_{\mathcal G(\tilde{\mD})},\mathbf I^{(m)}\big)\Big],
  \qquad
  \ell(\hat{\mathbf I},\mathbf I)=\|\hat{\mathbf I}-\mathbf I\|_2^2+\lambda\,\mathrm{LPIPS}(\hat{\mathbf I},\mathbf I),
\end{equation}
with $\lambda=0.05$ and $\beta=0.9$. $\mathcal G(\mD)$ is the Gaussian set built from the final depth and $\mathcal G(\tilde{\mD})$ the one built from the coarse depth of \cref{sec:prelim} through the same anchors, 3D stage, and head (shared weights); no depth ground truth is used. All models train for 150k steps with AdamW under \volsplat's training protocol (\cref{sec:appendix_impl,sec:appendix_protocol}). Nothing in \cref{sec:anchors,sec:theory,sec:head} has a data-dependent shape, so at inference the entire encoder runs as one fp16 CUDA graph (\cref{sec:realtime}).

\section{Experiments}
\label{sec:experiments}

\begin{table}[t]
  \centering
  \caption{\textbf{RealEstate10K~\citep{zhou2018stereo}, 6/12/24 views.} pixelSplat--GGN are quoted from \volsplat's retrainings~\citep{wang2025volsplat}; our retraining uses \volsplat's official protocol (\cref{sec:setup}). $^\ddagger$Pose-free, released checkpoints with ground-truth cameras (different training corpora and resolutions, no retraining). OOM: cannot infer on 96\,GB.}
  \label{tab:re10k_multiview}
  {\tabfont
  \begin{tabular}{l|ccc|ccc|ccc}
    \toprule
    \multirow{2}{*}{Method}
    & \multicolumn{3}{c|}{6 views} & \multicolumn{3}{c|}{12 views} & \multicolumn{3}{c}{24 views} \\
    \cmidrule(lr){2-4} \cmidrule(lr){5-7} \cmidrule(lr){8-10}
    & PSNR$\uparrow$ & SSIM$\uparrow$ & LPIPS$\downarrow$ & PSNR$\uparrow$ & SSIM$\uparrow$ & LPIPS$\downarrow$ & PSNR$\uparrow$ & SSIM$\uparrow$ & LPIPS$\downarrow$ \\
    \midrule
    pixelSplat~\citep{charatan2024pixelsplat} & 28.95 & 0.900 & 0.163 & OOM & OOM & OOM & OOM & OOM & OOM \\
    MVSplat~\citep{chen2024mvsplat}           & 29.13 & 0.924 & 0.091 & 26.97 & 0.912 & 0.101 & 26.23 & 0.903 & 0.108 \\
    TranSplat~\citep{kim2025transplat}        & 29.62 & 0.928 & 0.084 & 28.00 & 0.920 & 0.089 & 26.65 & 0.884 & 0.115 \\
    DepthSplat~\citep{xu2025depthsplat}       & 30.52 & 0.931 & 0.079 & 28.54 & 0.919 & 0.088 & 26.26 & 0.880 & 0.115 \\
    GGN~\citep{zhang2024gaussian}             & 26.68 & 0.879 & 0.133 & 25.83 & 0.870 & 0.142 & 23.86 & 0.840 & 0.171 \\
    \midrule
    AnySplat~\citep{jiang2025anysplat}$^\ddagger$    & 25.37 & 0.811 & 0.140 & 25.55 & 0.815 & 0.136 & 27.23 & 0.882 & 0.107 \\
    WorldMirror~\citep{liu2025worldmirror}$^\ddagger$ & 29.20 & 0.914 & 0.106 & 29.56 & 0.921 & 0.101 & 29.37 & 0.920 & 0.103 \\
    \midrule
    \volsplat (reported)                      & 31.30 & 0.941 & 0.075 & 29.40 & 0.928 & 0.085 & 27.21 & 0.896 & 0.111 \\
    \volsplat (retrained)                     & 30.93 & 0.938 & 0.078 & 30.86 & 0.939 & 0.077 & 27.24 & 0.881 & 0.127 \\
    \textbf{\ours (ours)}                     & \textbf{31.51} & \textbf{0.942} & \textbf{0.074} & \textbf{31.47} & \textbf{0.942} & \textbf{0.073} & \textbf{30.57} & \textbf{0.934} & \textbf{0.082} \\
    \bottomrule
  \end{tabular}
  }
\end{table}

\subsection{Experimental Setup}
\label{sec:setup}

\paragraph{Datasets and protocol.}
We train on RealEstate10K~\citep{zhou2018stereo} ($256\times256$) and evaluate on \volsplat's official indices ($\sim$700 test scenes, 8 target views each) at 6, 12, and 24 context views; zero-shot generalization uses ACID~\citep{liu2021infinite} (173 scenes), and indoor transfer follows the FreeSplat protocol~\citep{wang2024freesplat} on ScanNet~\citep{dai2017scannet}, fine-tuning every RealEstate10K checkpoint for 150k steps and evaluating on \volsplat's official 72-sequence index. All our models and our retrained \volsplat follow \volsplat's official training protocol (\cref{sec:training,sec:appendix_protocol}; $4\times$ NVIDIA RTX PRO 6000).

\paragraph{Baselines and metrics.}
We compare against pixel-aligned (pixelSplat, MVSplat, TranSplat, DepthSplat), post-processed (GGN), and pose-free (AnySplat, WorldMirror; released checkpoints) methods, FreeSplat on ScanNet, and \volsplat as published and retrained, the anchor for all paired analyses (\cref{sec:appendix_protocol}). We report PSNR, SSIM~\citep{wang2004image}, and LPIPS~\citep{zhang2018unreasonable}; all methods are timed with their released code on the same GPU under the same protocol (\cref{sec:appendix_efficiency}).

\subsection{Main Results on RealEstate10K}
\label{sec:main_results}

\begin{figure}[t]
  \centering
  \includegraphics[width=1.0\textwidth]{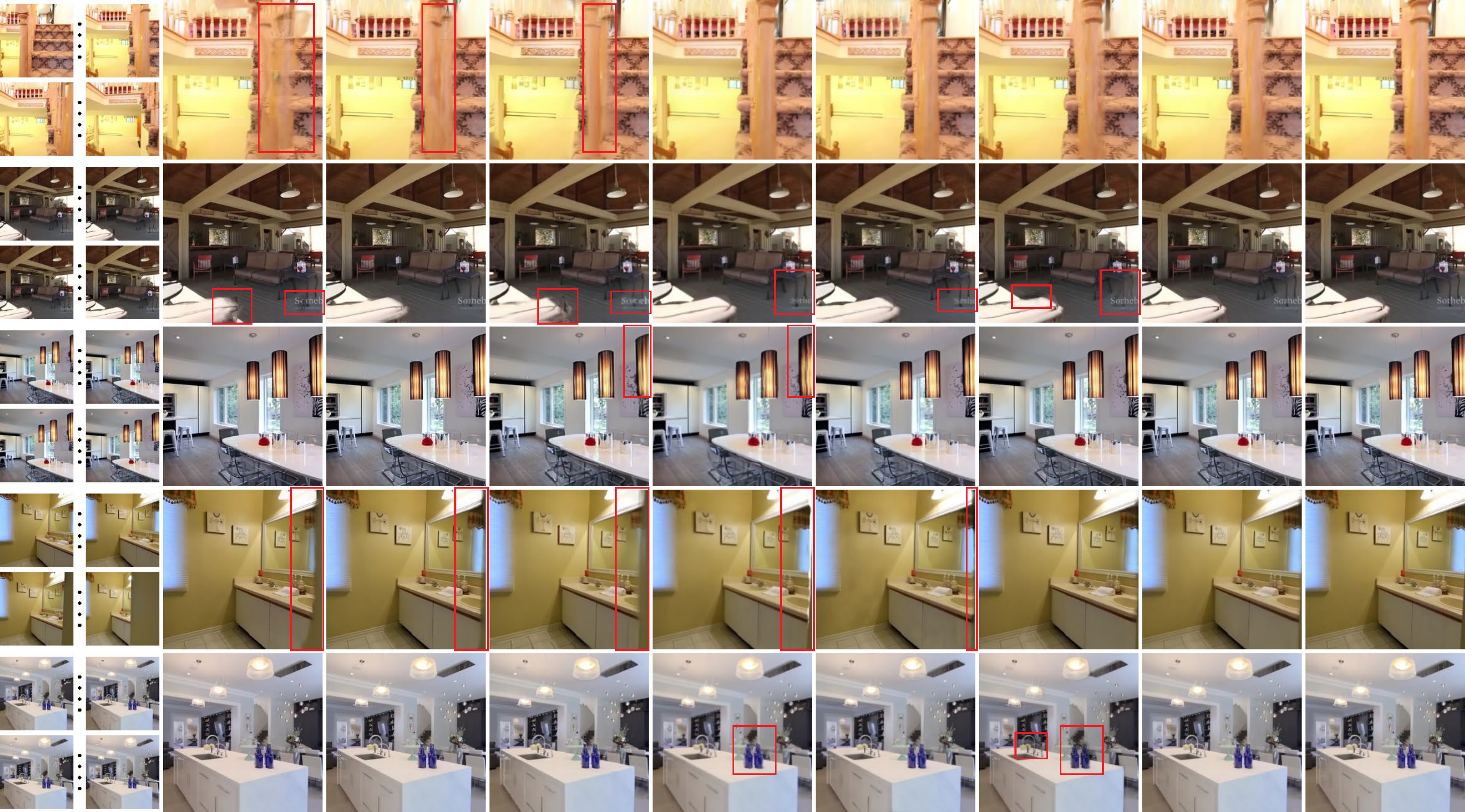}\\[1pt]
  \newcommand{\figcol}[1]{\makebox[0.10884\textwidth][c]{\footnotesize #1}}%
  \newcommand{\figgap}{\hspace{0.00255\textwidth}}%
  \figcol{Inputs}\figgap
  \figcol{MVSplat}\figgap
  \figcol{DepthSplat}\figgap
  \figcol{GGN}\figgap
  \figcol{AnySplat}\figgap
  \figcol{WorldMirror}\figgap
  \figcol{VolSplat}\figgap
  \figcol{Ours}\figgap
  \figcol{Ground Truth}
  \vspace{-0.25cm}
  \caption{\textbf{RealEstate10K~\citep{zhou2018stereo}, 6 input views.} \ours (``Ours'') preserves fine detail and thin structures with less ghosting; red boxes highlight local differences. Checkpoints follow \cref{tab:re10k_multiview} (\volsplat: our retraining; AnySplat and WorldMirror: released checkpoints with ground-truth cameras).}
  \label{fig:vis_re10k}
\end{figure}

\cref{tab:re10k_multiview} reports quality across view counts. \ours has the best PSNR, SSIM, and LPIPS at every view count: $+0.58$ PSNR over our retrained \volsplat at 6 views and $+0.61$ at 12 (paired, 68\% of scenes). Our retrained \volsplat is the stronger 12-view baseline (30.86 vs.\ the published 29.40) and anchors every paired comparison (\cref{sec:appendix_protocol}). At 24 views every RealEstate10K-trained method degrades as the context densifies, \volsplat most; \ours does not share the drop: 30.57 PSNR, $+3.3$ paired over \volsplat (98\% of scenes) and $+1.2$ over WorldMirror, which is pretrained on a large multi-dataset corpus and essentially flat in the view count; \cref{sec:ablations} traces the difference to one hyper-parameter of the depth branch. Qualitatively, pixel-aligned methods blur and duplicate thin structures at 6 views (\cref{fig:vis_re10k}); at 24 views \volsplat streaks and the pose-free baselines blur branches and text (\cref{fig:vis_re10k_24v}; 12 views in \cref{fig:vis_re10k_12v}).

\begin{table}[t]
  \centering
  \caption{\textbf{Six-view reconstruction efficiency} (RealEstate10K; same GPU and timing protocol for all methods, \cref{sec:appendix_efficiency}). Baselines run in plain fp32 as released; \ours on its deployed path (fp16, CUDA graph) and in plain fp32. Methods as in \cref{tab:re10k_multiview}. Training: 4 GPUs, batch 1/GPU, wall-clock extrapolated from the measured step rate to 150k steps. $^\ddagger$Pose-free, released defaults, not retrainable here.}
  \label{tab:efficiency}
  {\tabfont
  \begin{tabular}{lccccccc}
    \toprule
    Method & PSNR$\uparrow$ & \makecell{Params\\train.\ / total} & \makecell{Inference\\(ms)} & \makecell{Recon.\\FPS} & \makecell{Peak mem.\\(MB)} & \makecell{Train\\(it/s)} & \makecell{Train\\wall-clock} \\
    \midrule
    pixelSplat & 28.95 & 119.0 / 119.0\,M & 238.5 & 4.2 & 31{,}869 & 1.05 & $\sim$39.6\,h \\
    MVSplat           & 29.13 & 12.0 / 12.0\,M   & 93.8  & 10.7 & 3{,}528 & 2.92 & $\sim$14.3\,h \\
    DepthSplat       & 30.52 & 120.7 / 120.7\,M & 81.2  & 12.3 & 7{,}204 & 2.44 & $\sim$17.1\,h \\
    GGN             & 26.68 & 12.1 / 12.1\,M   & 177.0 & 5.6  & 3{,}489 & 2.25 & $\sim$18.5\,h \\
    \midrule
    AnySplat$^\ddagger$       & 25.37 & 886.4 / 1{,}190.7\,M   & 85.5 & 11.7 & 3{,}774 & --- & --- \\
    WorldMirror$^\ddagger$   & 29.20 & 1{,}263.3 / 1{,}263.3\,M & 90.8 & 11.0 & 5{,}673 & --- & --- \\
    \midrule
    \volsplat (retrained), fp32 & 30.93 & 51.4 / 138.0\,M & 137.0 & 7.3 & 7{,}271 & 1.18 & $\sim$35.4\,h \\
    \ours, fp32 eager & \textbf{31.51} & 12.5 / 99.1\,M & 56.9 & 17.6 & 3{,}873 & \textbf{3.17} & $\sim$\textbf{13.2\,h} \\
    \textbf{\ours, deployed} & \textbf{31.51} & 12.5 / 99.1\,M & \textbf{33.0} & \textbf{30.3} & \textbf{1{,}081} & \textbf{3.17} & $\sim$\textbf{13.2\,h} \\
    \bottomrule
  \end{tabular}
  }
\end{table}

\subsection{Real-Time Reconstruction}
\label{sec:realtime}

\cref{tab:efficiency} summarizes efficiency at 6 views. The 3D stage has no data-dependent shape (a sort-based voxel group-by padded to a static $V{\cdot}H{\cdot}W$ row count, fixed-table gathers and scatters, pointwise $g$ and $h$), so the whole encoder is captured as one fp16 CUDA graph (\cref{sec:appendix_impl}). \ours reconstructs $4.2\times$ faster than \volsplat on this path and $2.4\times$ faster at equal fp32 precision (56.9 vs.\ 137.0\,ms), trains $2.7\times$ faster (13.2 vs.\ 35.4\,h; 27 vs.\ 159\,h at 24 views), and uses 85\% less peak inference memory with 76\% fewer trainable parameters; \volsplat gains only 1.4\% from the same optimizations, as its sparse U-Net resists mixed precision, static shapes, and graph capture. Most of the speedup is architectural: removing the U-Net takes the fp32 eager encoder from 137.0 to 94.4\,ms \emph{while improving quality} (\cref{tab:evolution,tab:speed_ladder}). Against the released baselines, \ours is the fastest at every view count, at the lowest memory and the highest PSNR ($2.5\times$ faster than DepthSplat, the next fastest; 2.9 vs.\ 27.3\,GB for \volsplat at 24 views; \cref{tab:speed_all,tab:speed_detail}); MVSplat and GGN match its trainable-parameter count but run $2.8$--$5.4\times$ slower and lose $2.4$ and $4.8$ PSNR.

\subsection{Generalization: Zero-Shot ACID and ScanNet Fine-Tuning}
\label{sec:acid}

\begin{figure}[t]
  \centering
  \includegraphics[width=1.0\textwidth]{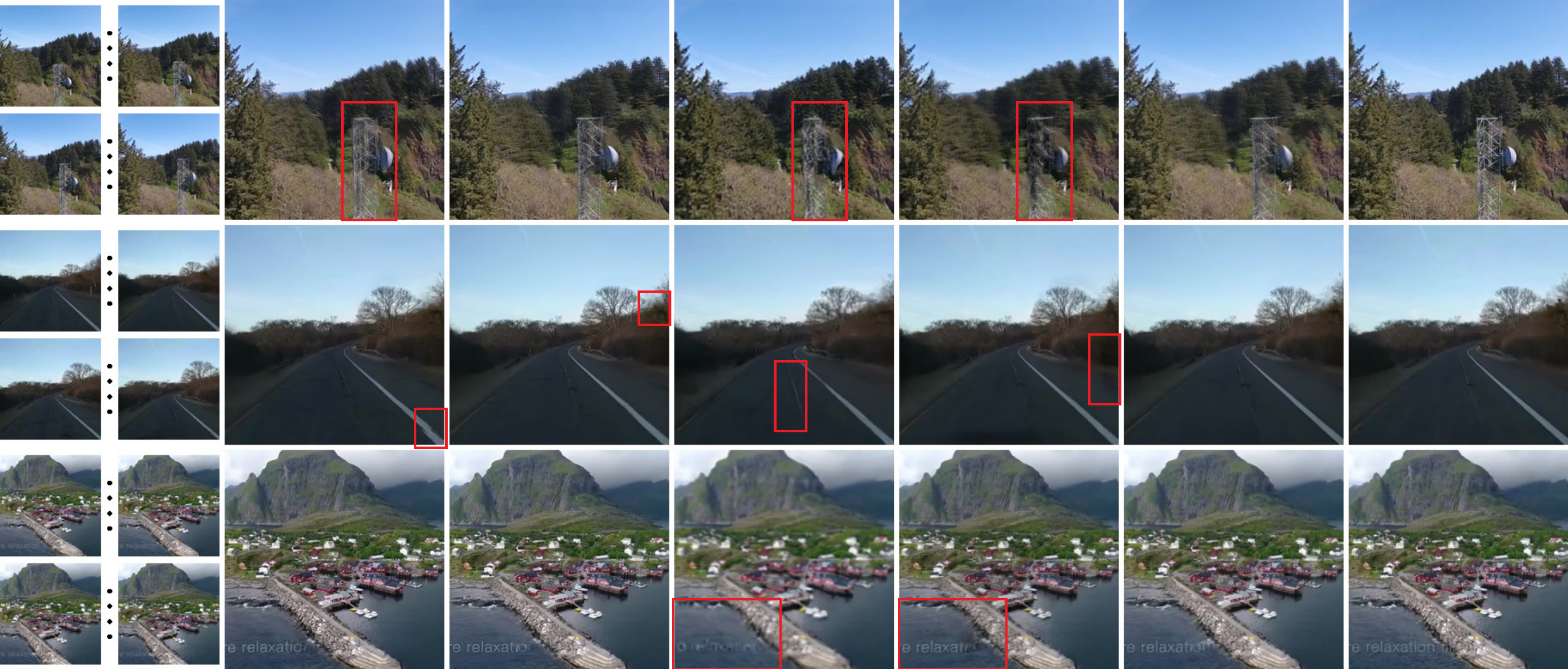}\\[1pt]
  \newcommand{\figcola}[1]{\makebox[0.14004\textwidth][c]{\footnotesize #1}}%
  \newcommand{\figgapa}{\hspace{0.00328\textwidth}}%
  \figcola{Inputs}\figgapa
  \figcola{MVSplat}\figgapa
  \figcola{DepthSplat}\figgapa
  \figcola{GGN}\figgapa
  \figcola{VolSplat}\figgapa
  \figcola{Ours}\figgapa
  \figcola{Ground Truth}
  \vspace{-0.25cm}
  \caption{\textbf{Zero-shot ACID~\citep{liu2021infinite}, 6 input views.} Models trained on RealEstate10K, checkpoints as in \cref{tab:acid}; \ours preserves thin structures and road markings with less blur and ghosting (red boxes).}
  \label{fig:vis_acid}
\end{figure}

\begin{figure}[t]
  \centering
  \includegraphics[width=1.0\textwidth]{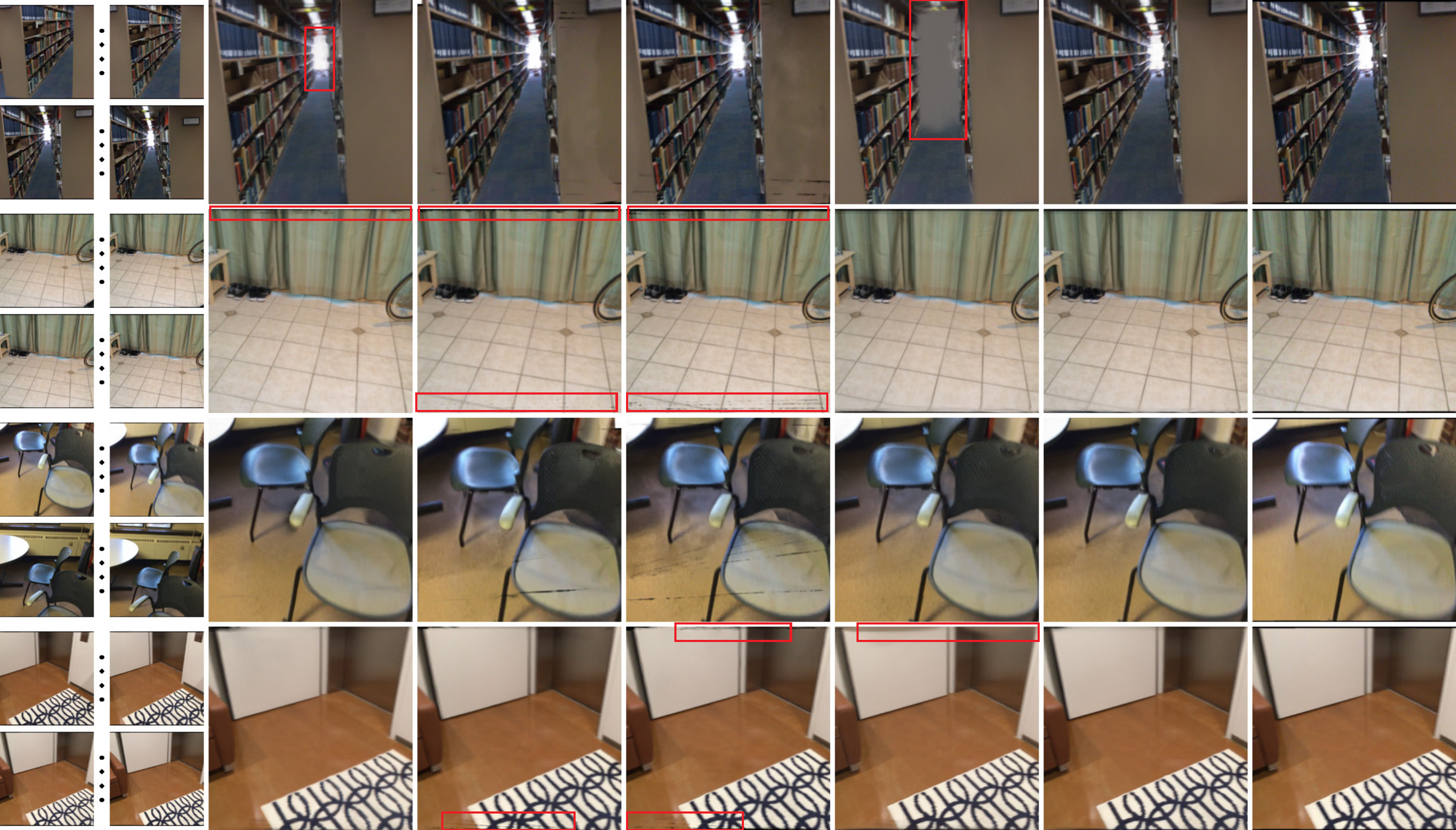}\\[1pt]
  \newcommand{\figcols}[1]{\makebox[0.14004\textwidth][c]{\footnotesize #1}}%
  \newcommand{\figgaps}{\hspace{0.00328\textwidth}}%
  \figcols{Inputs}\figgaps
  \figcols{FreeSplat}\figgaps
  \figcols{AnySplat}\figgaps
  \figcols{WorldMirror}\figgaps
  \figcols{VolSplat}\figgaps
  \figcols{Ours}\figgaps
  \figcols{Ground Truth}
  \vspace{-0.25cm}
  \caption{\textbf{ScanNet~\citep{dai2017scannet}, 6 input views.} \ours and \volsplat are fine-tuned under \volsplat's protocol, the other methods use released checkpoints (\cref{tab:scannet}); \ours preserves shelves and rug patterns with fewer artifacts.}
  \label{fig:vis_scannet}
\end{figure}

\cref{tab:acid} evaluates RealEstate10K-trained models on ACID without fine-tuning. The margin over \volsplat does not shrink out of domain: it grows to $+0.87$ PSNR paired (77\% of scenes) against $+0.58$ in domain, and \ours also exceeds the PSNR and LPIPS \volsplat reports for its own checkpoint. In \cref{fig:vis_acid}, \volsplat's refinement smears thin structures such as a radio tower and the pixel-aligned baselines ghost or dissolve them, which \ours preserves.
\label{sec:scannet}
\cref{tab:scannet} fine-tunes \ours and \volsplat from their RealEstate10K checkpoints under \volsplat's ScanNet fine-tuning protocol; the ordering carries over with a wider margin: \ours gains $+0.92$ PSNR over the fine-tuned \volsplat (on all 8 scenes) and lies above every published entry (\cref{fig:vis_scannet}, \cref{sec:appendix_vis_scannet}).

\begin{table}[t]
  \centering
  \begin{minipage}[t]{0.49\linewidth}
    \centering
    \caption{\textbf{Zero-shot RE10K$\to$ACID~\citep{liu2021infinite}, 6 views.} MVSplat, DepthSplat, and GGN: \volsplat's 6-view releases; only \volsplat (reported) is quoted, other rows are measured under its official protocol.}
        \label{tab:acid}
        {\tabfont
        \begin{tabular}{lccc}
          \toprule
          Method & PSNR$\uparrow$ & SSIM$\uparrow$ & LPIPS$\downarrow$ \\
          \midrule
          MVSplat    & 30.88 & 0.911 & 0.105 \\
          DepthSplat & 31.93 & 0.917 & 0.097 \\
          GGN        & 27.45 & 0.828 & 0.188 \\
          \midrule
          \volsplat (reported)  & 32.65 & \textbf{0.932} & 0.092 \\
          \volsplat (retrained) & 32.29 & 0.924 & 0.097 \\
          \textbf{\ours}   & \textbf{33.16} & 0.931 & \textbf{0.091} \\
          \bottomrule
        \end{tabular}
        }
  \end{minipage}
  \hfill
  \begin{minipage}[t]{0.49\linewidth}
    \centering
    \caption{\textbf{ScanNet~\citep{dai2017scannet}, 6 input views} (72 sequences; a \emph{fine-tuning} experiment). Baselines and \volsplat (reported) quoted from \volsplat~\citep{wang2025volsplat}; the last two rows use its 150k-step fine-tuning protocol; paired analysis in \cref{sec:acid}.}
      \label{tab:scannet}
      {\tabfont
      \begin{tabular}{lccc}
        \toprule
        Method & PSNR$\uparrow$ & SSIM$\uparrow$ & LPIPS$\downarrow$ \\
        \midrule
        FreeSplat   & 27.45 & 0.829 & 0.222 \\
        FreeSplat++ & 27.45 & 0.829 & 0.223 \\
        AnySplat    & 19.45 & 0.626 & 0.344 \\
        WorldMirror & 25.83 & 0.819 & 0.136 \\
        \midrule
        \volsplat (reported)  & 28.41 & 0.906 & 0.127 \\
        \volsplat (retrained) & 28.24 & 0.905 & 0.122 \\
        \textbf{\ours}   & \textbf{29.16} & \textbf{0.911} & \textbf{0.115} \\
        \bottomrule
      \end{tabular}
      }
  \end{minipage}
\end{table}

\begin{table}[t]
  \centering
  \caption{\textbf{Plane-sweep candidate count and the dense-view collapse.} Full-budget retrainings of \ours with one component changed. $\Delta u$: adjacent-candidate disparity relative to 6 views, $D{=}128$ (\cref{eq:disparity}); paired $\Delta$PSNR vs.\ \ours at the same view count. $D{=}256$ and $D{=}32$ were pre-registered predictions; the 12-view reference is re-evaluated for this table (31.49 vs.\ 31.47 in \cref{tab:re10k_multiview}).}
  \label{tab:ablation_rt24v}
  {\tabfont
  \begin{tabular}{lcc@{\quad}ccc@{\quad}cc}
    \toprule
    Model & $D$ & $\Delta u$ (rel.) & PSNR$\uparrow$ & SSIM$\uparrow$ & LPIPS$\downarrow$ & \makecell{Paired $\Delta$PSNR\\vs.\ \ours} & \makecell{\ours\\better} \\
    \midrule
    \multicolumn{8}{@{}l}{\emph{12 views}} \\
    \ours, $D{=}256$ (pre-reg.) & 256 & 0.23 & 28.39 & 0.895 & 0.109 & $-3.10$ & 99\% \\
    \textbf{\ours} & 64 & 0.92 & \textbf{31.49} & \textbf{0.943} & \textbf{0.073} & --- & --- \\
    \midrule
    \multicolumn{8}{@{}l}{\emph{24 views}} \\
    \volsplat (retrained) & 128 & 0.22 & 27.24 & 0.881 & 0.127 & $-3.3$ & 98\% \\
    \ours, $D{=}128$ & 128 & 0.22 & 28.08 & 0.888 & 0.114 & $-2.49$ & 98\% \\
    \ours, trilinear fine level & 64 & 0.45 & 30.45 & 0.932 & 0.084 & $-0.12$ & 65\% \\
    \ours, $D{=}32$ (pre-reg.) & 32 & 0.91 & 30.39 & 0.932 & 0.083 & $-0.18$ & 55\% \\
    \textbf{\ours} & 64 & 0.45 & \textbf{30.56} & \textbf{0.934} & \textbf{0.082} & --- & --- \\
    \bottomrule
  \end{tabular}
  }
\end{table}

\subsection{Ablations and Analysis}
\label{sec:ablations}

\paragraph{Statistical protocol.}
All ablations retrain under \volsplat's exact 150k-step training protocol with identical losses and seeds, changing one component at a time. Gaps between our own models are per-scene \emph{paired} differences; a gap within the replicate noise of repeated training runs ($\le$0.03 PSNR, \cref{sec:appendix_protocol}) is reported as a tie. Full ablation tables are in \cref{sec:appendix_ablation_tables}.

\paragraph{Ablation models.}
\label{sec:collapse}
The coarse-source ablations run on the \emph{two-grid model}: \volsplat's pipeline with the two grids and fusion MLP of \cref{sec:theory} as its 3D stage (trilinear splat and query, 0.16M-parameter $g$), keeping its 164-d features, sparse-convolutional head, and $D{=}128$ (\cref{tab:ablation_source}). The level-count controls (\cref{tab:ablation_levels_150k}) run on the final \ours configuration.

\begin{wrapfigure}{r}{0.5\linewidth}
  \centering
  \includegraphics[width=\linewidth]{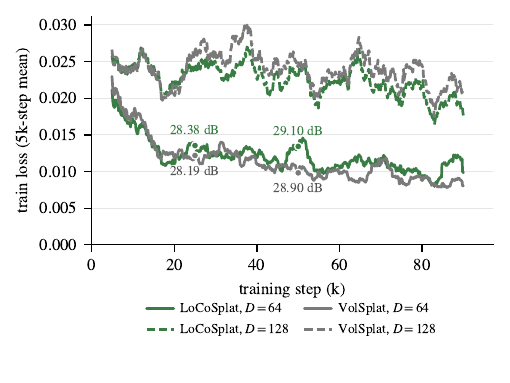}
  \caption{\textbf{24-view training loss} (5k-step mean). Both $D{=}128$ models plateau at twice the loss; the original \volsplat with $D{=}64$ follows \ours. Dots: PSNR on the 24-view index at 25k and 50k steps. Curves cover the first 90k steps of the 150k-step schedule.}
  \label{fig:loss_24v}
\end{wrapfigure}
\paragraph{Removing the U-Net helps under a fixed training protocol.}
\label{sec:ngp}
Without the U-Net, fixed grids fused by $g$ \emph{improve} PSNR by $+0.71$ with eight levels (31.64 vs.\ 30.93, 6 views) and $+0.70$ with the two levels of \ours, with 33\% fewer trainable parameters and $1.67\times$ higher training throughput (\cref{tab:evolution}). The gap holds at every matched checkpoint and widens with training: the U-Net's test PSNR plateaus after $\sim$80k steps while the grid model keeps improving. The 3D stage still needs only minimal context: dropping the coarse level costs \ours $0.13$ PSNR (\cref{tab:ablation_levels_150k}), four times the replicate noise. At 12 views the full pipeline gains $+0.61$ over our retrained \volsplat (paired, 68\% of scenes; \cref{tab:ablation_unet}).

\paragraph{The coarse context needs neither learning nor appearance.}
In place of the learned coarse level (\cref{tab:ablation_source}), (i)~a \emph{fixed, untrained} low-pass of the fine field (box pooling or Gaussian smoothing) ties it on all three metrics, and (ii)~a detached 10-channel geometric descriptor with no appearance content (density, position offset, per-axis spread, log-depth moments; \cref{sec:appendix_impl}) matches it ($+0.03$, a tie). On \ours, two levels tie four levels within $[0.001,0.008]$ and eight reaching $0.128$; only a wider window adds ($+0.12$ at $2.048$, \cref{tab:ablation_levels_150k}).

\paragraph{The plane-sweep candidate count, not the 3D stage, decides the 24-view drop.}
Retraining \ours at 24 views with one change reverted at a time (\cref{tab:ablation_rt24v}), trilinear fine splatting costs $0.1$ PSNR, whereas \volsplat's $D{=}128$ plane-sweep candidates (instead of 64) reproduce its failure (28.08 PSNR, worse on 98\% of scenes) with every 3D-stage component in place; every degrading 24-view model runs 128 candidates and every healthy one 64. Denser official indices shrink the matching baseline $B$, and the per-candidate disparity $\Delta u\propto B/(D-1)$ of the failing configuration is half that of the healthy ones (\cref{sec:appendix_24v}). Collapsed models have a switched-off multi-view depth head (the same one-hot candidate at every pixel, depth carried by the monocular path), whereas healthy ones are peaked and input-dependent (\cref{tab:mechanism}). The account passed both pre-registered tests: $D{=}256$ at 12 views ($\Delta u=0.23$) collapses ($-3.10$ paired, 99\% of scenes) and $D{=}32$ at 24 views ($\Delta u=0.91$) stays healthy ($-0.18$). The converse holds on the baseline itself: retraining the original \volsplat at 24 views with $D{=}64$ and nothing else changed puts it on the healthy loss trajectory of \ours (\cref{fig:loss_24v}), so the drop is a property of the depth branch's candidate spacing, not of the 3D stage.

\section{Related Work}
\label{sec:related_work}

\paragraph{Per-scene novel view synthesis.}
NeRF~\citep{mildenhall2021nerf} and its successors~\citep{barron2021mip,fridovich2022plenoxels,muller2022instant,barron2023zip} and 3D Gaussian Splatting~\citep{kerbl20233d,lu2024scaffold,fan2024lightgaussian,ren2024octree} optimize each scene; we target the feed-forward regime, where reconstruction itself must be fast.

\paragraph{Feed-forward 3DGS.}
Generalizable NVS networks~\citep{yu2021pixelnerf,wang2021ibrnet,chen2021mvsnerf} amortize reconstruction across scenes; pixelSplat~\citep{charatan2024pixelsplat} and MVSplat~\citep{chen2024mvsplat} predict pixel-aligned Gaussians from epipolar or cost-volume features; TranSplat~\citep{kim2025transplat}, DepthSplat~\citep{xu2025depthsplat}, and FreeSplat~\citep{wang2024freesplat,wang2025freesplat} improve depth estimation and cross-view fusion; GGN~\citep{zhang2024gaussian} and AnySplat~\citep{jiang2025anysplat} post-process pixel-aligned Gaussians, and WorldMirror~\citep{liu2025worldmirror} predicts them jointly with geometry. A pose-free line~\citep{ye2024no,huang2025spfsplat}, to which AnySplat and WorldMirror also belong, builds on DUSt3R- and VGGT-style geometry transformers~\citep{wang2024dust3r,wang2025vggt} pretrained on large multi-dataset corpora, while SelfSplat~\citep{kang2025selfsplat} does so without 3D priors; these models change the inputs and the training data rather than the 3D stage, so we report them as released, for reference. \volsplat~\citep{wang2025volsplat} unprojects per-pixel features along predicted depth, voxelizes them, refines the sparse volume with a 3D U-Net~\citep{cciccek20163d,choy20194d}, and decodes one Gaussian per occupied voxel; related systems likewise lean on sparse 3D CNNs~\citep{zhang2022nerfusion} or volumetric fusion~\citep{miao2025evolsplat}, while Long-LRM~\citep{ziwen2024long} instead scales a long-sequence Mamba/transformer model over image tokens.

\paragraph{Multi-resolution grids and efficient inference.}
Instant-NGP~\citep{muller2022instant} popularized multi-resolution feature grids with tiny MLPs for per-scene optimization, on a long line of grid and octree structures~\citep{meagher1982geometric,riegler2017octnet,koneputugodage2023octree,fridovich2023k}. Our grids store no learned scene content and are rebuilt from the input images at every forward pass: they are a splat-and-query aggregation structure that computes binned local means of point features, in the sense of binned kernel estimators~\citep{hall1996binned}, not a learned representation.
\section{Conclusion}
\label{sec:conclusion}

We have presented \ours, a real-time feed-forward 3DGS method whose 3D stage has no learned 3D network, only a fixed average that a grid computes for free. Its premise is that a Gaussian is local: once depth is known, the 3D stage has only to supply the surface layout around each anchor. Controlled ablations show that this 3D stage needs very little: a fixed low-pass context read by a tiny pointwise map beats the sparse 3D U-Net under an identical training protocol and trains faster, and the minimal coarse context that remains useful needs neither learning nor appearance content. \ours outperforms \volsplat on every metric at 6, 12, and 24 views, in and out of domain, and runs as a single fp16 CUDA graph at 30 FPS, $4.2\times$ faster than \volsplat and the fastest of seven feed-forward methods. A second finding lies outside the 3D stage: for this pipeline the 24-view collapse is decided by the plane-sweep candidate count, and a disparity-resolution account of it predicted, before training, both a new collapse and a healthy dense-view model.

\paragraph{Limitations.}
The 3D-stage ablations run only at 6 views and only against \volsplat's U-Net, not smaller 3D CNNs or local attention, and we have not evaluated beyond the posed $256\times256$ regime: at $512^2$, where $H=0.128$ may not transfer, on wider-baseline data such as DL3DV, or under noisy depth or pose error. In principle the geometry is depth-dominated: the head moves each centre at most $1.5v$ from its anchor (\cref{eq:head}), so the 3D layout inherits the quality of the depth prediction, and the 24-view collapse indeed occurs in the depth branch. The plain mean is optimal only for a smooth field with homoscedastic noise (\cref{sec:appendix_theory}) and blurs across surfaces at occlusion boundaries, and the pointwise head relies on anchor isolation (\cref{rem:pointwise}), which denser inputs may erode.

\subsection*{AI use statement}

We used LLM-based assistants in three ways. First, for writing: to
draft sections of the paper from the authors' notes, results, and
outlines, to polish the text, and to check compliance with the
conference template. Second, for retrieval and discovery: to help find
related work. Third, as coding assistants: part of the implementation,
evaluation, and plotting code was generated by LLM-based tools under
the authors' supervision. They were not used to generate
research ideas, hypotheses, theory, proofs, or methods, and every
claim and number in the paper originates from the authors' own work.
All drafted or edited text and all generated code were reviewed by the
authors, who take full responsibility for the content of this paper.
\subsection*{Ethics statement}

We use only public research datasets (RealEstate10K, ACID, ScanNet) under their licenses; no new data were collected and no human subjects were involved. Our contribution lowers the cost of existing feed-forward reconstruction rather than enabling new capabilities, and carries the same dual-use considerations as prior work in the area. We are not aware of any conflicts of interest.

\subsection*{Reproducibility statement}

All models, including our retrained \volsplat baseline, follow \volsplat's public training protocol and official evaluation indices (\cref{sec:setup}). Architecture and optimizer details are in \cref{sec:appendix_impl}, the timing protocol and hardware in \cref{sec:appendix_efficiency}, the paired-comparison protocol in \cref{sec:ablations,sec:appendix_protocol}, and the selection procedure for qualitative figures in \cref{sec:appendix_vis12v,sec:appendix_vis24v}. Code, configurations, evaluation indices, and checkpoints will be released upon publication.

\bibliography{references}
\bibliographystyle{iclr2027_conference}

\appendix
\crefalias{section}{appendix}\crefalias{subsection}{subappendix} 
\crefname{appendix}{Appendix}{Appendices}\Crefname{appendix}{Appendix}{Appendices}
\crefname{subappendix}{Appendix}{Appendices}\Crefname{subappendix}{Appendix}{Appendices}
\section*{Appendix}

\etocdepthtag.toc{appendix}
\etocsettagdepth{main}{none}
\etocsettagdepth{appendix}{subsection}
\etocsettocstyle{}{}
\tableofcontents

\section{Efficiency Measurements}
\label{sec:appendix_efficiency}

\subsection{Benchmark Protocol}
All timings (\cref{tab:efficiency,tab:speed_all,tab:speed_detail}) are measured on a single idle RTX PRO 6000 (96\,GB; PyTorch 2.8, CUDA 12.8) with one common protocol: each method alone on the GPU, the same 100 RealEstate10K test scenes, 10 warm-up scenes, batch size 1, $256^2$ inputs, 8 target views, per-scene medians of synchronized wall-clock time. ``Encoder'' maps context images (plus cameras) to Gaussians; ``render'' rasterizes all 8 targets with the method's own renderer. FPS is $1000/t_{\mathrm{enc}}$ (or $1000/(t_{\mathrm{enc}}+t_{\mathrm{render}})$); memory is peak allocated. Baselines run as released (fp32; AnySplat/WorldMirror with a bf16 ViT at $252^2$, WorldMirror with ground-truth camera priors); \ours runs its deployed path (\cref{sec:appendix_impl}), and its fp32 eager path gives the same ordering. Checkpoints: the 6-view releases of the \volsplat authors for pixelSplat/MVSplat/DepthSplat/GGN (their cost is shape-only, so reuse at 12/24 views leaves timing unchanged; pixelSplat's is 6-view-specific), authors' weights for AnySplat/WorldMirror, and our retrained \volsplat and \ours. Code changes were limited to dependency fallbacks and two one-line fixes. TranSplat is not timed (its code hard-codes 2--4 views). ``Trainable'' parameters are those each method's training protocol updates; ``total'' adds frozen backbones.

\subsection{Training Throughput}
All six retrainable methods run their own code under our training protocol (4 GPUs, DDP, batch 1/GPU, 6 context + 8 target views, MSE + LPIPS), timed over steps 50--350 and extrapolated to 150k steps (MVSplat/GGN received pixelSplat's multi-view sampler). \ours trains at 3.17 steps/s vs.\ 1.18 for \volsplat, with 20\,GB peak memory per GPU vs.\ 37 (\volsplat), 66 (pixelSplat), and 23--28\,GB (others); our full 150k run including validation took 17.6\,h. AnySplat and WorldMirror fine-tune billion-parameter backbones on dataset mixtures and cannot be retrained under this protocol.

\begin{table}[h]
  \centering
  \caption{\textbf{Reconstruction speed, 6/12/24 views} (RealEstate10K, 100-scene medians; protocol above). FPS counts scenes; ``+ render'' includes 8 targets. Checkpoints: MVSplat, DepthSplat, and GGN are timed at 12 and 24 views with their released 6-view checkpoints (their cost is shape-only, so timing is unaffected); \volsplat at 24 views with our 12-view checkpoint ($^\S$); \ours with its own 6-, 12-, and 24-view checkpoints. Baselines run in plain fp32 as released, \ours on its deployed fp16 CUDA-graph path (fp32 eager: 56.9\,ms at 6 views, the same ordering). $^\ddagger$Pose-free, released defaults: $252^2$, bf16 backbone, multi-dataset pretraining; WorldMirror uses ground-truth camera priors.}
  \label{tab:speed_all}
  {\tabfont
  \begin{tabular}{l|ccc|ccc|ccc}
    \toprule
    \multirow{2}{*}{Method} & \multicolumn{3}{c|}{Encoder (ms) $\downarrow$} & \multicolumn{3}{c|}{Encoder-only FPS $\uparrow$} & \multicolumn{3}{c}{Encoder + render, FPS $\uparrow$} \\
    \cmidrule(lr){2-4} \cmidrule(lr){5-7} \cmidrule(lr){8-10}
    & 6v & 12v & 24v & 6v & 12v & 24v & 6v & 12v & 24v \\
    \midrule
    pixelSplat  & 238.5 & ---   & ---    & 4.2  & --- & --- & 4.0  & --- & --- \\
    MVSplat            & 93.8  & 303.9 & 1111.7 & 10.7 & 3.3 & 0.9 & 9.6  & 3.1 & 0.9 \\
    DepthSplat        & 81.2  & 149.8 & 306.4  & 12.3 & 6.7 & 3.3 & 11.1 & 6.1 & 3.0 \\
    GGN              & 177.0 & 483.8 & 1451.7 & 5.6  & 2.1 & 0.7 & 5.4  & 2.0 & 0.7 \\
    AnySplat$^\ddagger$     & 85.5  & 116.6 & 209.0  & 11.7 & 8.6 & 4.8 & 10.5 & 7.6 & 4.3 \\
    WorldMirror$^\ddagger$ & 90.8  & 142.0 & 259.0  & 11.0 & 7.0 & 3.9 & 10.7 & 6.8 & 3.7 \\
    \volsplat (retrained) & 137.0 & 263.8 & 548.5$^\S$  & 7.3  & 3.8 & 1.8$^\S$ & 6.9  & 3.6 & 1.7$^\S$ \\
    \midrule
    \textbf{\ours} (deployed)                 & \textbf{33.0} & \textbf{62.9} & \textbf{137.4} & \textbf{30.3} & \textbf{15.9} & \textbf{7.3} & \textbf{23.7} & \textbf{12.9} & \textbf{6.1} \\
    \bottomrule
  \end{tabular}
  }
\end{table}

\begin{table}[h]
  \centering
  \caption{\textbf{Rendering, memory, and Gaussian counts.} Protocol and $^\ddagger$ as in \cref{tab:speed_all}.}
  \label{tab:speed_detail}
  {\tabfont
  \begin{tabular}{l|ccc|ccc|ccc}
    \toprule
    \multirow{2}{*}{Method} & \multicolumn{3}{c|}{Render 8 views (ms) $\downarrow$} & \multicolumn{3}{c|}{Peak memory (MB) $\downarrow$} & \multicolumn{3}{c}{\#Gaussians (mean)} \\
    \cmidrule(lr){2-4} \cmidrule(lr){5-7} \cmidrule(lr){8-10}
    & 6v & 12v & 24v & 6v & 12v & 24v & 6v & 12v & 24v \\
    \midrule
    pixelSplat              & 10.4 & ---  & ---  & 31{,}869 & ---     & ---      & 393{,}216 & ---       & --- \\
    MVSplat                 & 9.9  & 17.2 & 31.7 & 3{,}528 & 7{,}277  & 24{,}341 & 393{,}216 & 786{,}432 & 1{,}572{,}864 \\
    DepthSplat              & 8.4  & 13.4 & 23.8 & 7{,}204 & 13{,}912 & 27{,}330 & 393{,}216 & 786{,}432 & 1{,}572{,}864 \\
    GGN                     & 9.2  & 9.3  & 9.8  & 3{,}489 & 7{,}158  & 24{,}010 & 176{,}837 & 238{,}445 & 319{,}885 \\
    AnySplat$^\ddagger$     & 10.4 & 14.9 & 23.8 & 3{,}774 & 4{,}199  & 5{,}106  & 343{,}347 & 640{,}670 & 1{,}145{,}219 \\
    WorldMirror$^\ddagger$  & 2.9  & 5.0  & 8.7  & 5{,}673 & 6{,}497  & 8{,}152  & 326{,}098 & 577{,}272 & 972{,}238 \\
    \volsplat (retrained)   & 8.6  & 14.6 & 25.5 & 7{,}271 & 13{,}964 & 27{,}349 & 393{,}175 & 786{,}288 & 1{,}572{,}369 \\
    \midrule
    \ours                   & 9.1  & 14.8 & 26.5 & \textbf{1{,}081} & \textbf{1{,}695} & \textbf{2{,}925} & 393{,}216 & 786{,}432 & 1{,}572{,}864 \\
    \bottomrule
  \end{tabular}
  }
\end{table}

\subsection{Scaling and Speedup Decomposition}
\paragraph{Scaling.}
Deployed \ours is fastest at every view count, and the ordering follows how views are aggregated: pixelSplat's epipolar transformer is slowest already at 6 views; MVSplat and GGN attend across view pairs and scale near-quadratically ($11.9\times$ / $8.2\times$ from 6 to 24 views); DepthSplat and \volsplat scale linearly from a high constant; the pose-free models scale sub-linearly but start $2.6$--$2.7\times$ slower. \ours scales linearly from the lowest constant (its $4.2\times$ from 6 to 24 views reflects the static stage padded to $V{\cdot}256^2$ rows) and is the only encoder under 150\,ms at 24 views, with 1.1/1.7/2.9\,GB peak memory vs.\ $\ge$24\,GB for every pixel-aligned baseline at 24 views. \volsplat's time varies most across scenes (s.d.\ 20--40\,ms), since its sparse U-Net cost follows voxel occupancy.

\paragraph{Speedup decomposition.}
With no dynamic sparse computation left, the whole encoder is one static-shape graph, so every systems optimization applies end-to-end. Of the $4.2\times$ speedup (\cref{tab:evolution,tab:speed_ladder}), the architecture accounts for most: dropping the U-Net and pruning to two levels (137.0$\to$94.4\,ms, \emph{while improving quality}), the graph-native retrain (58.9\,ms), and two backbone-side levers (56.9\,ms in fp32, still $+0.58$ PSNR over \volsplat). fp16, type-check removal, whole-encoder CUDA graphs, and stream parallelism give the rest (33.0\,ms; 42\,ms including 8 rendered views). Checkpoint-compatible steps are metric-identical ($\le10^{-4}$ PSNR); the retrain steps cost $31.63\to31.51$ in total. The same optimizations speed \volsplat up by only 1.4\%: its dominant sparse U-Net benefits from none of these optimizations.

\subsection{Training at High View Counts}
Without recomputation, 24-view \volsplat needs $\sim$126\,GB of fp32 activations. With exact activation recomputation (deltas $\le6\times10^{-6}$) its 24-view run peaked at 53\,GB and took 159\,h, with per-GPU memory growing from 48 to 62\,GB, versus 27\,h at flat memory for \ours-24v.

\begin{table}[h]
  \centering
  \caption{\textbf{Cumulative speedups.} Top: checkpoint-compatible optimizations; bottom: retrained models with deployed stacks.}
  \label{tab:speed_ladder}
  {\tabfont
  \begin{tabular}{lcc}
    \toprule
    Configuration & Encoder (ms) & PSNR \\
    \midrule
    Two-grid model (\cref{sec:collapse}), original code & 94.4 & 31.63 \\
    \;+ fused ops (SDPA, packed voxel unique, fused gathers) & 87.4 & 31.63 \\
    \;+ fp16 image-side autocast & 83.8 & 31.63 \\
    \;+ runtime type-checking removed & 74.4 & 31.63 \\
    \;+ CUDA-graphed 2D image side & 67.4 & 31.63 \\
    \midrule
    graph-native retrain (pointwise head, fine-from-base, $D{=}64$) & 52.0 & 31.60 \\
    \;+ stream parallelism (mono $\parallel$ multi-view) & 47.0 & 31.60 \\
    \;+ ViT-B backbone, ViT-S-width interface & 37.7 & 31.58 \\
    \;+ DWConv half-width regressor (= \ours, full stack) & \textbf{33.0} & 31.51 \\
    \bottomrule
  \end{tabular}
  }
\end{table}

\section{Implementation and Protocol Details}
\subsection{Implementation Details}
\label{sec:appendix_impl}

\begin{figure}[t]
  \centering
  \includegraphics[width=1.0\textwidth]{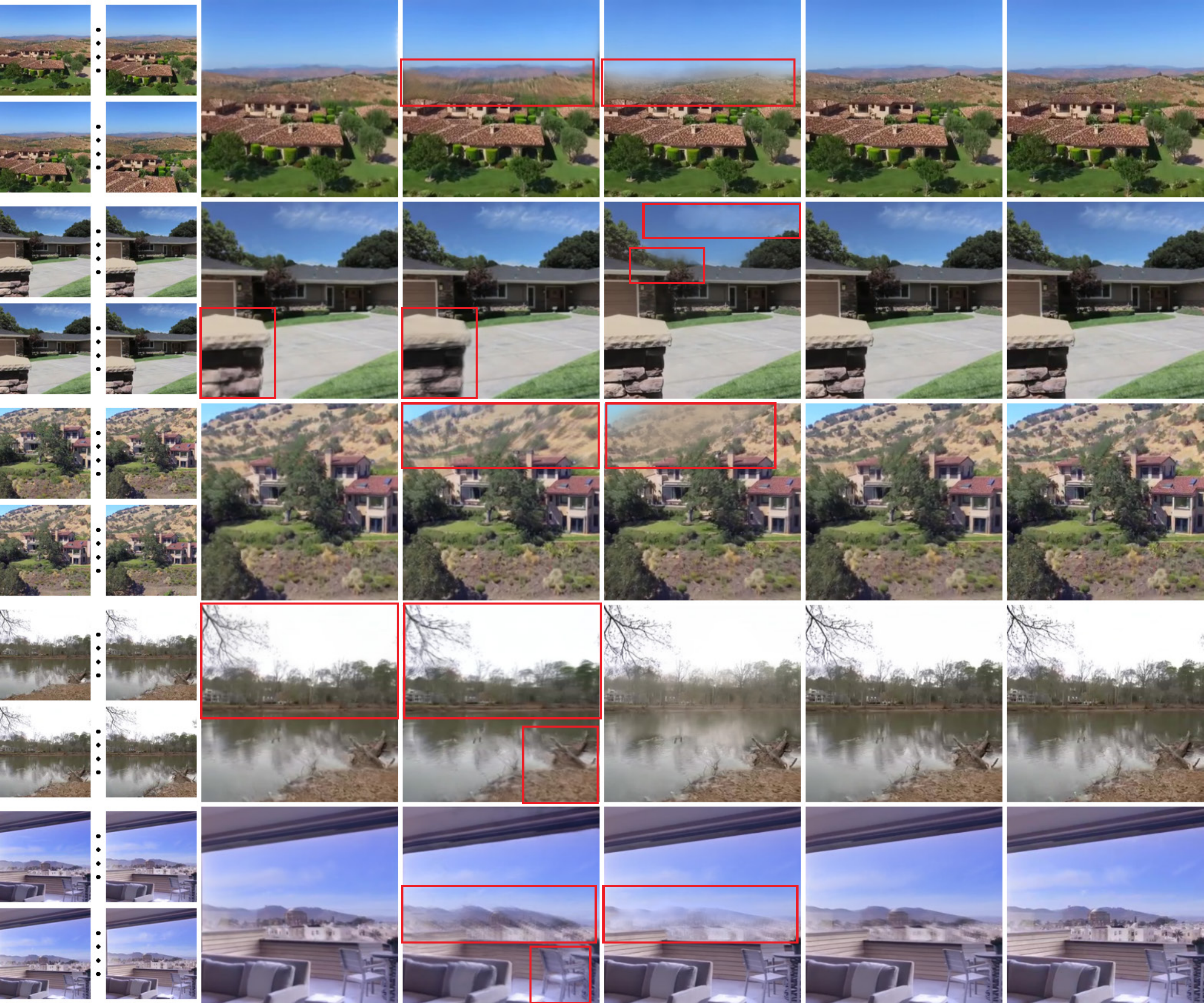}\\[1pt]
  \newcommand{\figcolb}[1]{\makebox[0.16347\textwidth][c]{\footnotesize #1}}%
  \newcommand{\figgapb}{\hspace{0.00383\textwidth}}%
  \figcolb{Inputs}\figgapb
  \figcolb{AnySplat}\figgapb
  \figcolb{WorldMirror}\figgapb
  \figcolb{VolSplat}\figgapb
  \figcolb{Ours}\figgapb
  \figcolb{Ground Truth}
  \vspace{-0.25cm}
  \caption{\textbf{RealEstate10K~\citep{zhou2018stereo}, 12 input views} (four shown). \ours (``Ours'') preserves masonry, branches, and distant structures where baselines blur or introduce floaters. Red boxes highlight local differences. \volsplat is retrained at 12 views; pose-free baselines use released checkpoints with ground-truth cameras, as in \cref{fig:vis_re10k}.}
  \label{fig:vis_re10k_12v}
\end{figure}

\paragraph{Grids.}
Both grids store the feature mean $\vu_H$ (\cref{eq:nw}) of a linear projection of the point features; no density, position, or covariance channel. For points $(\vx_i,\vf_i)$, $\vf_i\in\mathbb R^{132}$ (164 for the ablation models of \cref{sec:collapse}), and levels $\ell\in\{0,1\}$ with $h_0=0.001$, $h_1=0.128$:
\begin{enumerate}
  \item \emph{Project}: $\mW_\ell\in\mathbb R^{16\times132}$ per level (4.3k parameters, the only learned part of the grids).
  \item \emph{Splat}: each point goes to its nearest node $c_\ell(\vx_i)=\operatorname{rnd}(\vx_i/h_\ell)$ (box kernel) and each occupied cell stores the mean of $\mW_\ell\vf_i$; the ablation models use weight-normalized trilinear splatting (tent kernel). Empty cells hold zero.
  \item \emph{Query}: anchor $a$ reads $\vu_\ell(c_\ell(\vx_a))$ (ablation models: trilinear, renormalized over occupied corners). Since $h_0=v$, the fine level equals the per-voxel mean $\mW_0\vf_a$ exactly and is taken directly from the base voxelization.
  \item \emph{Fuse}: $\vf'_a=\vf_a+g([\vu_0;\vu_1;\vf_a])$, with $g$ an MLP $164\to256\to256\to132$ (0.14M).
\end{enumerate}
Nearest-cell splat/query is quality-neutral against trilinear. The geometry-only coarse level of \cref{tab:ablation_source} (two-grid model only, not part of \ours) replaces steps~2--3 at the coarse level with a detached 10-channel descriptor, zero-padded to 16: offset of the cell's mean point from the anchor over $H$ (3), per-axis s.d.\ of the positions in the cell over $H$ (3), mean and s.d.\ of log depth (2), $\log(1+n_H)/5$ with $n_H$ the cell's point count (1), and query occupancy (1).

\paragraph{Head and parameters.}
The two-grid model keeps \volsplat's two $3^3$ sparse convolutions ($164\to38\to38$, 0.21M); \ours uses a pointwise MLP ($132\to38\to38$, 6.5k). Trainable / total parameters (86.6M of them the frozen ViT-B): \volsplat 51.4 / 138.0M, two-grid model 34.5 / 121.1M, \ours 12.5 / 99.1M.

\paragraph{Deployed inference stack.}
An exact sort-based voxel group-by padded to a static $B\!\cdot\!V\!\cdot\!H\!\cdot\!W$ row count (dead rows get zero opacity and are moved out of view) removes all data-dependent shapes and host syncs, so the entire encoder is captured as one CUDA graph. fp16 autocast covers the image side, with depth values and pose/warp math pinned to fp32 and the 3D stage in fp32 (68.8\,dB agreement with fp32 renders; benchmark metrics identical to four decimals). Further steps: SDPA with cached masks, fused cost-volume gathers, no runtime type-checking, and stream parallelism between the monocular and multi-view branches.

\paragraph{Backbone-side levers.}
A ViT-S-width interface to the retained ViT-B backbone (four trainable $768{\to}384$ projections) recovers $\sim$88\% of the quality a full ViT-S swap loses, so the pretrained prior, not the width, is what matters. A depthwise-separable half-width cost-volume regressor is quality-neutral, whereas a plain-conv width cut loses 0.45 PSNR.

\subsection{Experimental Protocol}
\label{sec:appendix_protocol}

\begin{figure}[t]
  \centering
  \includegraphics[width=1.0\textwidth]{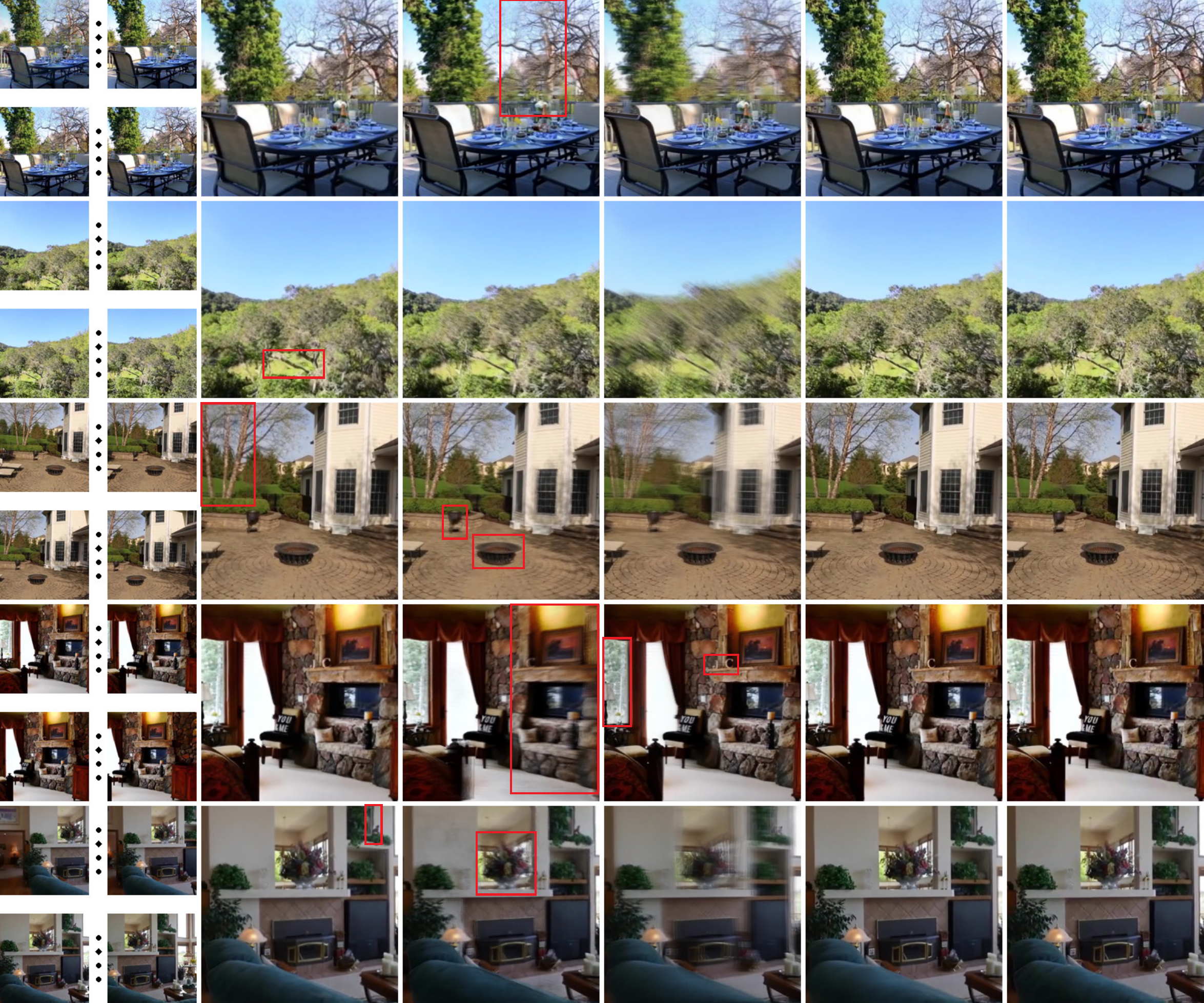}\\[1pt]
  \newcommand{\figcold}[1]{\makebox[0.16347\textwidth][c]{\footnotesize #1}}%
  \newcommand{\figgapd}{\hspace{0.00383\textwidth}}%
  \figcold{Inputs}\figgapd
  \figcold{AnySplat}\figgapd
  \figcold{WorldMirror}\figgapd
  \figcold{VolSplat}\figgapd
  \figcold{Ours}\figgapd
  \figcold{Ground Truth}
  \vspace{-0.25cm}
  \caption{\textbf{RealEstate10K~\citep{zhou2018stereo}, 24 input views} (four shown). In these selected examples, \ours (``Ours'') preserves branches, text, and fine structure where \volsplat streaks and pose-free baselines blur. Red boxes highlight local differences. Checkpoints, evaluation paths, and example selection are detailed below.}
  \label{fig:vis_re10k_24v}
\end{figure}

\paragraph{Data and training.}
We train on RealEstate10K~\citep{zhou2018stereo} ($256^2$) and evaluate on \volsplat's official indices ($\sim$700 scenes, 8 targets) at 6/12/24 views, zero-shot on ACID~\citep{liu2021infinite} (173 scenes), and on ScanNet~\citep{dai2017scannet} following FreeSplat~\citep{wang2024freesplat}: each RealEstate10K checkpoint is fine-tuned for 150k steps with \volsplat's training protocol and evaluated on its 72-sequence index. All our models and our retrained \volsplat share this protocol: 150k steps, $4\times$ RTX PRO 6000, batch 1/GPU, AdamW~\citep{loshchilov2017decoupled} with a one-cycle schedule, lr $2\times10^{-4}$ ($2\times10^{-6}$ for the backbone), MSE + LPIPS + intermediate depth supervision, identical seeds; 12/24-view runs use exact activation recomputation.

\paragraph{Baselines.}
Pixel-aligned methods (pixelSplat~\citep{charatan2024pixelsplat}, MVSplat~\citep{chen2024mvsplat}, TranSplat~\citep{kim2025transplat}, DepthSplat~\citep{xu2025depthsplat}), GGN~\citep{zhang2024gaussian}, the pose-free AnySplat~\citep{jiang2025anysplat} and WorldMirror~\citep{liu2025worldmirror}, and FreeSplat~\citep{wang2024freesplat,wang2025freesplat} on ScanNet; numbers are quoted from \volsplat~\citep{wang2025volsplat} unless noted. Our retrained \volsplat anchors all paired analyses. It is within $0.4$ PSNR of the published model at 6 views (30.93 vs.\ 31.30) and 24 views (27.24 vs.\ 27.21) and stronger at 12 views (30.86 vs.\ 29.40), so every 12-view gain is measured against the retrained model.

\paragraph{Metrics and ties.}
PSNR, SSIM~\citep{wang2004image}, and LPIPS~\citep{zhang2018unreasonable}; efficiency as in \cref{sec:appendix_efficiency}. Comparisons between our models use per-scene paired differences. Three seed draws of \ours give 31.51 / 31.50 / 31.50 (max paired difference 0.02, s.d.\ $<0.01$), and three seed draws of the two-grid model lie pairwise within $0.03$ (s.d.\ $0.01$), so differences $\le$0.03 PSNR are reported as ties.

\section{Analysis of the 3D Stage}
\subsection{The Path to the Two-Grid Model}
\label{sec:appendix_discussion}

\begin{table}[t]
  \centering
\caption{\textbf{Cumulative evolution to \ours} (150k steps, RealEstate10K, 6 views, 709 scenes). Enc.: encoder time on one RTX PRO 6000; intermediate models run their original code, \ours its full inference stack (\cref{tab:speed_ladder}).}
    \label{tab:evolution}
    {\tabfont
    \begin{tabular}{lccc}
      \toprule
      Stage & PSNR$\uparrow$ & LPIPS$\downarrow$ & Enc.\ (ms) \\
      \midrule
      \volsplat (sparse 3D U-Net, 17M) & 30.93 & 0.078 & 137.0 \\
      \;$\to$ 8-level grid + MLP       & \textbf{31.64} & \textbf{0.073} & 118 \\
      \;$\to$ 2 levels $\{0.001,0.128\}$ & 31.63 & \textbf{0.072} & 94.4 \\
      \;$\to$ pointwise head, $D{=}64$ & 31.60 & 0.073 & 58.9 \\
      \;$\to$ ViT-B/narrow iface, DWConv (= \ours) & 31.51 & 0.074 & \textbf{33.0} \\
      \bottomrule
    \end{tabular}
    }
\end{table}

\paragraph{An eight-level grid without a 3D network.}
The starting hypothesis was that \volsplat's single-resolution voxelization loses both fine detail and context, and that an Instant-NGP-style~\citep{muller2022instant} multi-resolution grid, rebuilt from the images at every forward pass rather than learned, could keep both without a 3D network. Eight trilinear levels ($0.001\cdot2^\ell$) plus a 0.18M fusion MLP beat the 17M U-Net by $+0.71$ PSNR with 33\% fewer trainable parameters and $1.67\times$ training throughput; the gap held at every checkpoint and widened with training, while the U-Net's test PSNR plateaued after $\sim$80k steps.

\paragraph{The coarse scale, not the level count, drives the gain.}
Varying only the grid levels of \ours at full budget (\cref{tab:ablation_levels_150k}) shows that removing the coarse level is the one change that hurts ($-0.13$, $4\times$ replicate noise). Level count and coarsest scale within $[0.008,0.128]$ do not matter: 2, 4, and 8 levels tie (31.49--31.51), as 2 vs.\ 8 levels did on the two-grid model (31.63 vs.\ 31.64, at 20\% faster encoding). Only a \emph{wider} window helps: 12 levels reaching $2.048$ gain $+0.12$.

\paragraph{The coarse level is a fixed geometric low-pass.}
In \cref{tab:ablation_source}, untrained box pooling or Gaussian smoothing of the fine field ties the learned re-splat, so the context needs neither learning nor smooth interpolation (hence the nearest-cell grids of \ours). A geometry-only descriptor with no appearance content also ties ($+0.030$): low-order geometric moments of the cell carry the coarse context. Only the gridless radius mean trails ($-0.062$), a kernel-choice effect. Removing the fusion MLP costs $-0.044$; we keep it for its negligible cost.

\paragraph{The two-grid model.}
It keeps \volsplat's depth, features, anchors, head, and renderer, and its 3D stage is a fine $0.001$ grid, a coarse $0.128$ grid, and a 0.16M fusion MLP, with no U-Net; its only remaining sparse convolution is the retained head. It reaches 31.63 PSNR at 94.4\,ms (top row of \cref{tab:speed_ladder}); three training seeds lie pairwise within $0.03$ PSNR (\cref{sec:appendix_protocol}).

\subsection{Design Notes}
\label{sec:appendix_theory}

\begin{figure}[t]
  \centering
  \includegraphics[width=1.0\textwidth]{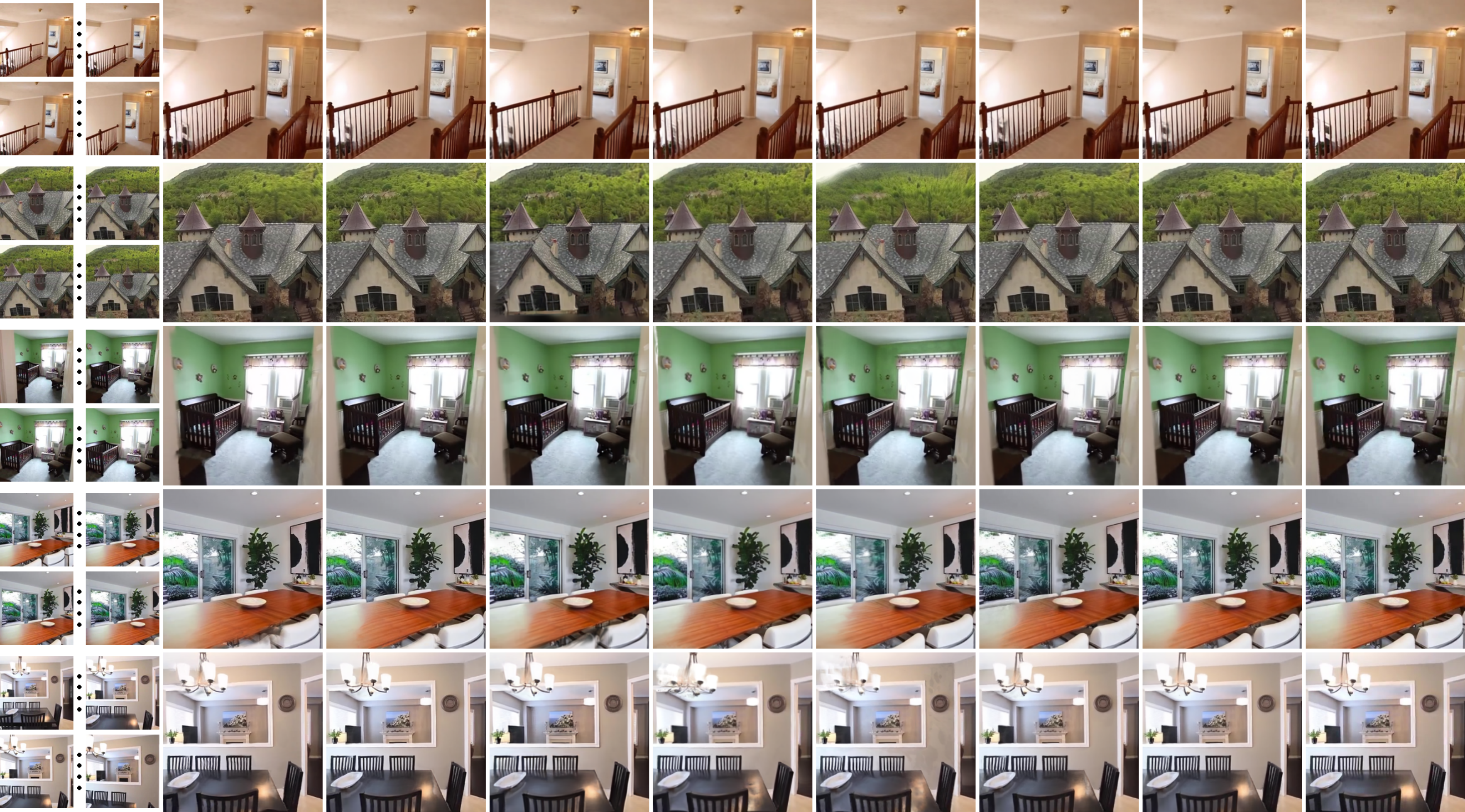}\\[1pt]
  \newcommand{\figcole}[1]{\makebox[0.10884\textwidth][c]{\footnotesize #1}}%
  \newcommand{\figgape}{\hspace{0.00255\textwidth}}%
  \figcole{Inputs}\figgape
  \figcole{MVSplat}\figgape
  \figcole{DepthSplat}\figgape
  \figcole{GGN}\figgape
  \figcole{AnySplat}\figgape
  \figcole{WorldMirror}\figgape
  \figcole{VolSplat}\figgape
  \figcole{Ours}\figgape
  \figcole{Ground Truth}
  \vspace{-0.25cm}
  \caption{\textbf{More qualitative comparisons on RealEstate10K~\citep{zhou2018stereo}, 6 input views} (four shown). Checkpoints and evaluation follow \cref{fig:vis_re10k}; selection is described in \cref{sec:appendix_vis_more}. Red boxes highlight local differences.}
  \label{fig:vis_re10k_more_6v}
\end{figure}

Box (nearest-cell) and tent (trilinear) grids correspond to the simple and linear binning rules of binned kernel estimators~\citep{hall1996binned}, here applied to \cref{eq:nw}.

\paragraph{Receptive field of the U-Net.}
Applying $\mathrm{RF}_{\mathrm{out}}=\mathrm{RF}_{\mathrm{in}}+(k-1)j$ through \volsplat's sparse U-Net (three stride-2 encoder blocks, two bottleneck blocks, four transposed-conv decoder stages) gives 93 fine voxels, i.e.\ $\pm0.046$ around each anchor, whereas a $0.128$ cell extends at least $\pm0.064$. The U-Net thus sees about a third of the context the coarse grid uses, and approximates the normalization by the cell's point count only through strided sums.

\paragraph{The plain mean as default aggregator.}
Under a smooth-field-plus-homoscedastic-noise model, the cell mean in \cref{eq:splat} is the minimum-variance unbiased linear estimate of the low-pass (Gauss--Markov). A reduced-budget pilot on the two-grid model (50k steps, one GPU, single seed) found no alternative aggregator that improved on it: confidence weighting by the detached matching probability hurt ($-0.23$ PSNR), and learned per-point weights, a per-cell variance channel, and a $3^3$-cell neighbourhood query all tied within the seed noise of that budget.

\begin{remark}[Sparse convolutions on isolated anchors]
\label{rem:pointwise}
A sparse convolution $(\mW\ast\vf)(a)=\sum_{o}\mW_o\vf_{a+o}\mathbf 1[a+o\in\mathcal A]$ reduces to its centre tap $\mW_0\vf_a$ whenever the $3^3$ neighbourhood of $a$ in the anchor set $\mathcal A$ is empty, so a stack of such layers interleaved with pointwise operations is a pointwise function at such anchors. Sparse-conv and pointwise heads therefore differ only on the fraction $p$ of anchors with an occupied neighbour, empirically $p=0.42\%$ at $v=0.001$ and 6 views (\cref{sec:anchors}). Hence \ours's pointwise head, which is quality-neutral at full budget (\cref{tab:evolution}).
\end{remark}

\subsection{Ablation Tables}
\label{sec:appendix_ablation_tables}

\cref{tab:ablation_levels_150k} (level count vs.\ scale range) and \cref{tab:ablation_source} (coarse source) are discussed in \cref{sec:appendix_discussion}; \cref{tab:ablation_unet} gives the paired 12-view comparison against our retrained \volsplat.

\begin{table}[htbp]
  \centering
  \caption{\textbf{\ours vs.\ \volsplat at 12 views.} RealEstate10K, 12 input views. Our retraining of \volsplat and \ours (which also carries the backbone-side changes of \cref{sec:prelim}) on the official 12-view index; $\Delta$ is the per-scene paired difference against the retrained \volsplat. The full \ours pipeline, whose 3D stage is the two-level grid and a pointwise MLP, gains $+0.61$, a whole-pipeline gain (the isolated 3D-stage comparison is \cref{tab:evolution} at 6 views). Our 12-view retraining (30.86) is stronger than the published model (29.40, \cref{sec:appendix_protocol}).}
  \label{tab:ablation_unet}
  {\tabfont
  \begin{tabular}{llcccc}
    \toprule
    Model (12 views) & 3D stage & PSNR$\uparrow$ & SSIM$\uparrow$ & LPIPS$\downarrow$ & $\Delta$PSNR \\
    \midrule
    \volsplat (retrained) & sparse 3D U-Net & 30.86 & 0.939 & 0.077 & --- \\
    \textbf{\ours} & two-level grid + MLP & \textbf{31.47} & \textbf{0.942} & \textbf{0.073} & $\mathbf{+0.61}$ vs.\ retrained (paired) \\
    \bottomrule
  \end{tabular}
  }
\end{table}

\begin{table}[htbp]
  \centering
  \caption{\textbf{Coarse-source ablation on the two-grid model} (\cref{sec:collapse}: \volsplat's 164-d features, trilinear grids, sparse-convolutional head, $D{=}128$; full 150k budget; RealEstate10K, 6 views, 709 scenes). Its reference is therefore the two-grid model (31.60 / 31.63), not \ours (31.51, \cref{tab:ablation_levels_150k}). The reference row (learned re-splat of the 16-d projected features) is the coarse level \ours ships. Paired $\Delta$PSNR vs.\ the reference run. The reference row lists two training runs at the same seed (this code tree / earlier run); three training seeds of the two-grid model lie pairwise within $0.03$ PSNR (\cref{sec:appendix_protocol}), which bounds the replicate noise (three draws of \ours give $0.02$, \cref{tab:ablation_levels_150k}), and differences within it are marked (tie).}
  \label{tab:ablation_source}
  {\tabfont
  \begin{tabular}{llcc}
    \toprule
    Coarse source & Learned? & PSNR$\uparrow$ & Paired $\Delta$ vs.\ ref.\ \\
    \midrule
    Learned re-splat (reference, 2 runs, same seed) & \checkmark & 31.60 / 31.63 & --- \\
    Fine field, hard box pooling & $\times$ & 31.58 & $-0.013$ (tie) \\
    Fine field, Gaussian low-pass & $\times$ & 31.59 & $-0.005$ (tie) \\
    Fine field, anchor-centered radius mean & $\times$ & 31.53 & $-0.062$ \\
    \textbf{Geometry-only statistics (no appearance)} & $\times$ & \textbf{31.63} & $+0.030$ (tie) \\
    \bottomrule
  \end{tabular}
  }
\end{table}

\begin{table}[htbp]
  \centering
  \caption{\textbf{Level count vs.\ scale range on the final \ours configuration at full budget} (only the grid levels change; RealEstate10K, 6 views, 709 scenes; the reference is \ours itself, 31.51, unlike \cref{tab:ablation_source}). $\Delta$PSNR is the difference of the 709-scene means vs.\ \ours, which on a common scene set equals the paired mean. (tie): within replicate noise, bounded by three seed draws of \ours (rows 2--4: largest paired difference $0.02$ PSNR, sample s.d.\ $<0.01$) and by three seed draws of the two-grid model of \cref{tab:ablation_source} ($\sim$0.03). Removing the coarse levels is the one change that hurts (1 level vs.\ 4 / 8 levels, paired: $-0.123$ / $-0.113$); four narrow levels, eight wide levels in two seed draws, and the two levels of \ours in three seed draws tie (8 vs.\ 4, paired: $-0.010$; 8 vs.\ 2 at the same seed: $-0.006$); extending the coarsest scale from $0.128$ to $2.048$ is the only change that helps.}
  \label{tab:ablation_levels_150k}
  {\tabfont
  \begin{tabular}{llcccc}
    \toprule
    Config & Scales & PSNR$\uparrow$ & SSIM$\uparrow$ & LPIPS$\downarrow$ & $\Delta$PSNR vs.\ \ours \\
    \midrule
    1 level (fine only) & $0.001$ & 31.38 & 0.941 & 0.075 & $-0.13$ \\
    \textbf{2 levels (\ours)} & $\{0.001, 0.128\}$ & 31.51 & 0.942 & 0.074 & --- \\
    2 levels (\ours, second seed) & $\{0.001, 0.128\}$ & 31.50 & 0.941 & 0.074 & $-0.01$ (tie) \\
    2 levels (\ours, third seed) & $\{0.001, 0.128\}$ & 31.50 & 0.941 & 0.074 & $-0.01$ (tie) \\
    4 levels ($\times2$) & $0.001$--$0.008$ & 31.50 & 0.941 & 0.074 & $-0.01$ (tie) \\
    8 levels ($\times2$) & $0.001$--$0.128$ & 31.49 & 0.941 & 0.074 & $-0.02$ (tie) \\
    8 levels ($\times2$, second seed) & $0.001$--$0.128$ & 31.50 & 0.942 & 0.074 & $-0.01$ (tie) \\
    12 levels ($\times2$) & $0.001$--$2.048$ & \textbf{31.62} & \textbf{0.942} & \textbf{0.073} & $\mathbf{+0.12}$ \\
    \bottomrule
  \end{tabular}
  }
\end{table}

\section{The 24-View Drop in Detail}
\label{sec:appendix_24v}

\begin{figure}[t]
  \centering
  \includegraphics[width=0.80\textwidth]{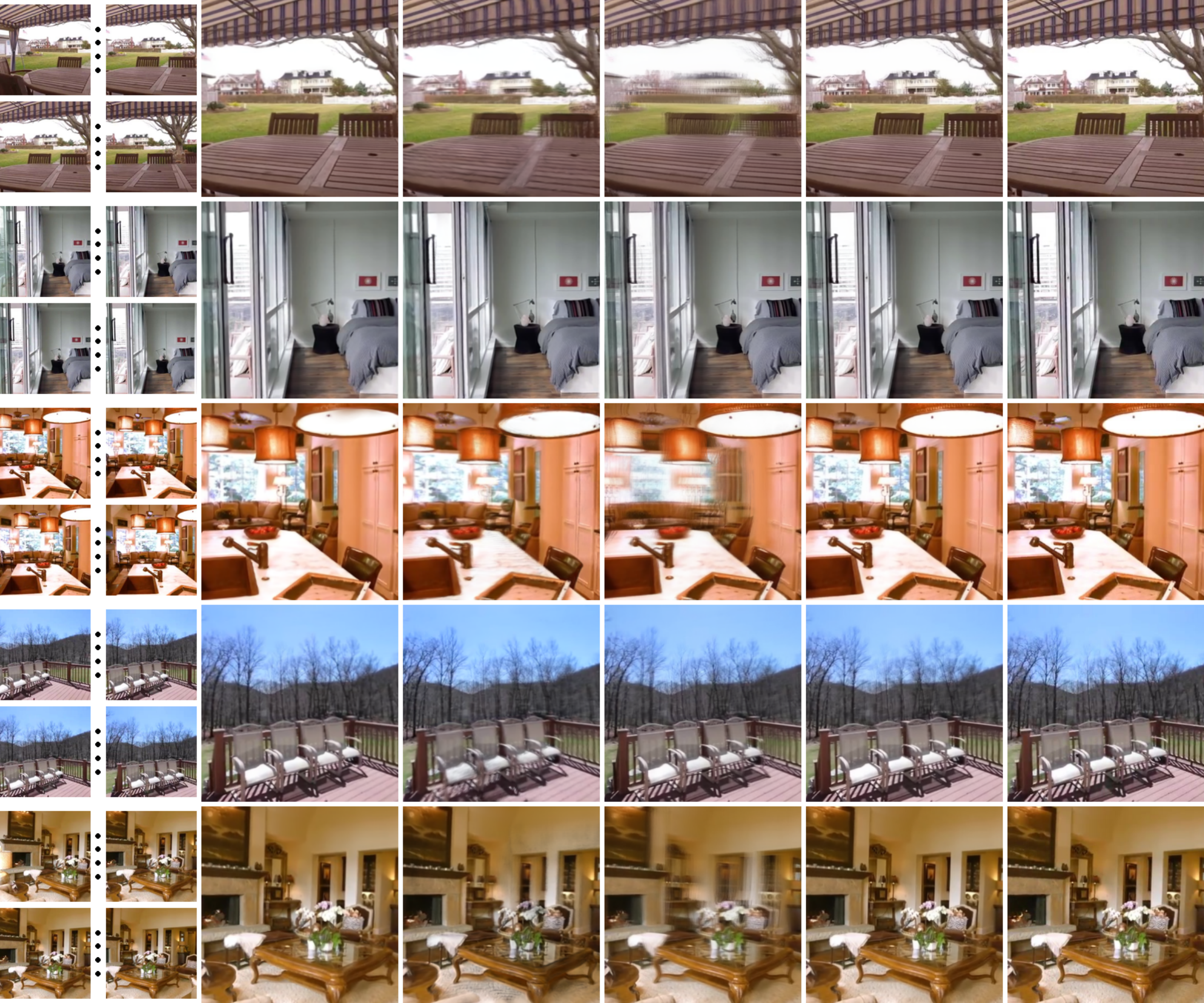}\\[1pt]
  \newcommand{\figcolg}[1]{\makebox[0.13078\textwidth][c]{\footnotesize #1}}%
  \newcommand{\figgapg}{\hspace{0.00306\textwidth}}%
  \figcolg{Inputs}\figgapg
  \figcolg{AnySplat}\figgapg
  \figcolg{WorldMirror}\figgapg
  \figcolg{VolSplat}\figgapg
  \figcolg{Ours}\figgapg
  \figcolg{Ground Truth}
  \vspace{-0.25cm}
  \caption{\textbf{More qualitative comparisons on RealEstate10K~\citep{zhou2018stereo}, 24 input views} (four shown). Checkpoints and evaluation follow \cref{fig:vis_re10k_24v}; selection is described in \cref{sec:appendix_vis_more}. Red boxes highlight local differences.}
  \label{fig:vis_re10k_more_24v}
\end{figure}

\subsection{The Drop Lies in the Depth Branch, Not the 3D Stage}
We retrained \ours at 24 views with one change reverted at a time (\cref{tab:ablation_rt24v}). Reverting the fine level to trilinear costs only 0.12 PSNR, but reverting the plane-sweep from $D{=}64$ to \volsplat's $D{=}128$ reproduces \volsplat's failure (28.08 vs.\ its 27.24 PSNR; $-2.49$ paired, worse on 98\% of scenes), with the same loss plateau at twice the healthy level from the first 15k steps, even with the grids and pointwise head in place. Every degrading 24-view model uses $D{=}128$ and every healthy one $D{=}64$: the drop is an optimization failure of the multi-view depth branch, which \ours avoided incidentally through a change made for speed.

\subsection{A Disparity-Resolution Account}
With $D$ candidates uniform in inverse depth on $[i_{\min},i_{\max}]$, adjacent candidates are separated in a source view at baseline $B$ by
\begin{equation}
\label{eq:disparity}
  \Delta u \;=\; f\,B\,\frac{i_{\max}-i_{\min}}{D-1},
\end{equation}
with $f$ the focal length in feature pixels. The official indices pack more views into a fixed $\sim$48-frame span, so the neighbour spacing (and with it $B$) falls from 9.5 frames at 6 views to 4.4 and 2.1. Relative to 6 views with $D{=}128$, $\Delta u$ is 0.46 (12v, $D{=}128$), 0.45 (24v, $D{=}64$), and 0.22 (24v, $D{=}128$): the two healthy dense settings share $\Delta u$, the failing one has half of it. We then pre-registered two predictions, both confirmed (\cref{tab:ablation_rt24v}): $D{=}256$ collapses the 12-view model ($\Delta u=0.23$; $-3.10$ paired PSNR, worse on 99\% of scenes, same loss signature), while $D{=}32$ keeps the 24-view model healthy ($\Delta u=0.91$; only $-0.18$, loss trajectory identical to $D{=}64$). Frame spacing is a faithful proxy: exact disparities from the evaluation cameras (0.21--0.91) match the spacing-based ones (0.22--0.92). $\Delta u$ thus predicts the collapse across all trained configurations.

\subsection{The Collapsed Depth Head}
Collapsed models have a \emph{switched-off} multi-view depth head (\cref{tab:mechanism}): all matching mass sits on one candidate, the same one at every pixel of every scene, so the head ignores its input, its depth is a constant 36--48\% off the final depth, and the monocular path carries depth. Healthy models are peaked and input-dependent (multi-view depth within 1.5--3.2\% of the final). Three facts point to optimization rather than a missing signal: (i) the absolute disparity is sub-pixel in all configurations (0.012--0.055 feature px), so no resolution threshold separates the groups, only the ratio does; (ii) the raw cost volume carries a matching signal everywhere, just spread over more bins when $\Delta u$ is small (90\%-mass width 55--74\% of the range vs.\ 20--49\%); (iii) at initialization all configurations are broad and alike, so the constant one-hot is learned.

\begin{table}[htbp]
  \centering
  \caption{\textbf{The depth branch of healthy and collapsed models.} $\Delta u$: exact median per-candidate disparity in scale-0 feature pixels over the full official index (709/696/680 scenes). Remaining columns: scale-0 matching distribution on the first 50 scenes of the index, pooled over all pixels: normalized entropy $H/\log D$ (median), $w_{90}$ = bins around the argmax holding 90\% of the mass (median), share of pixels whose argmax is the most frequent bin, and median relative error of the multi-view depth against the final fused depth. Collapsed models are input-independent one-hots at a fixed candidate; at initialization every configuration is broad and alike. Two-grid model: \cref{sec:collapse}.}
  \label{tab:mechanism}
  {\tabfont
  \begin{tabular}{lcccccccl}
    \toprule
    Model & Views & $D$ & \makecell{$\Delta u$\\(px)} & $H/\log D$ & \makecell{$w_{90}$\\(bins)} & \makecell{Modal-bin\\share} & \makecell{MV depth\\err.} & Status \\
    \midrule
    \ours & 6 & 64 & 0.110 & 0.42 & 7 & 13\% & 1.8\% & healthy \\
    \ours & 12 & 64 & 0.050 & 0.51 & 8 & 14\% & 2.0\% & healthy \\
    \ours, $D{=}256$ & 12 & 256 & 0.012 & 0.000 & 1 & 100\% & 48\% & collapsed \\
    \ours, $D{=}32$ & 24 & 32 & 0.047 & 0.51 & 5 & 19\% & 1.5\% & healthy \\
    \ours & 24 & 64 & 0.023 & 0.47 & 8 & 14\% & 3.2\% & healthy \\
    \ours, $D{=}128$ & 24 & 128 & 0.012 & 0.000 & 1 & 100\% & 46\% & collapsed \\
    \midrule
    Two-grid model & 6 & 128 & 0.055 & 0.39 & 8 & 7\% & 1.9\% & healthy \\
    Two-grid model & 12 & 128 & 0.025 & 0.46 & 14 & 10\% & 2.5\% & healthy \\
    Two-grid model & 24 & 128 & 0.012 & 0.000 & 1 & 100\% & 36\% & collapsed \\
    \midrule
    Any config.\ at init.\ & 6--24 & 32--256 & --- & 0.45--0.78 & 0.78--0.99\,$D$ & 50--99\% & --- & untrained \\
    \bottomrule
  \end{tabular}
  }
\end{table}

\section{Additional Qualitative Results}
\subsection{Qualitative Results at 12 Input Views}
\label{sec:appendix_vis12v}

\begin{figure}[t]
  \centering
  \includegraphics[width=0.80\textwidth]{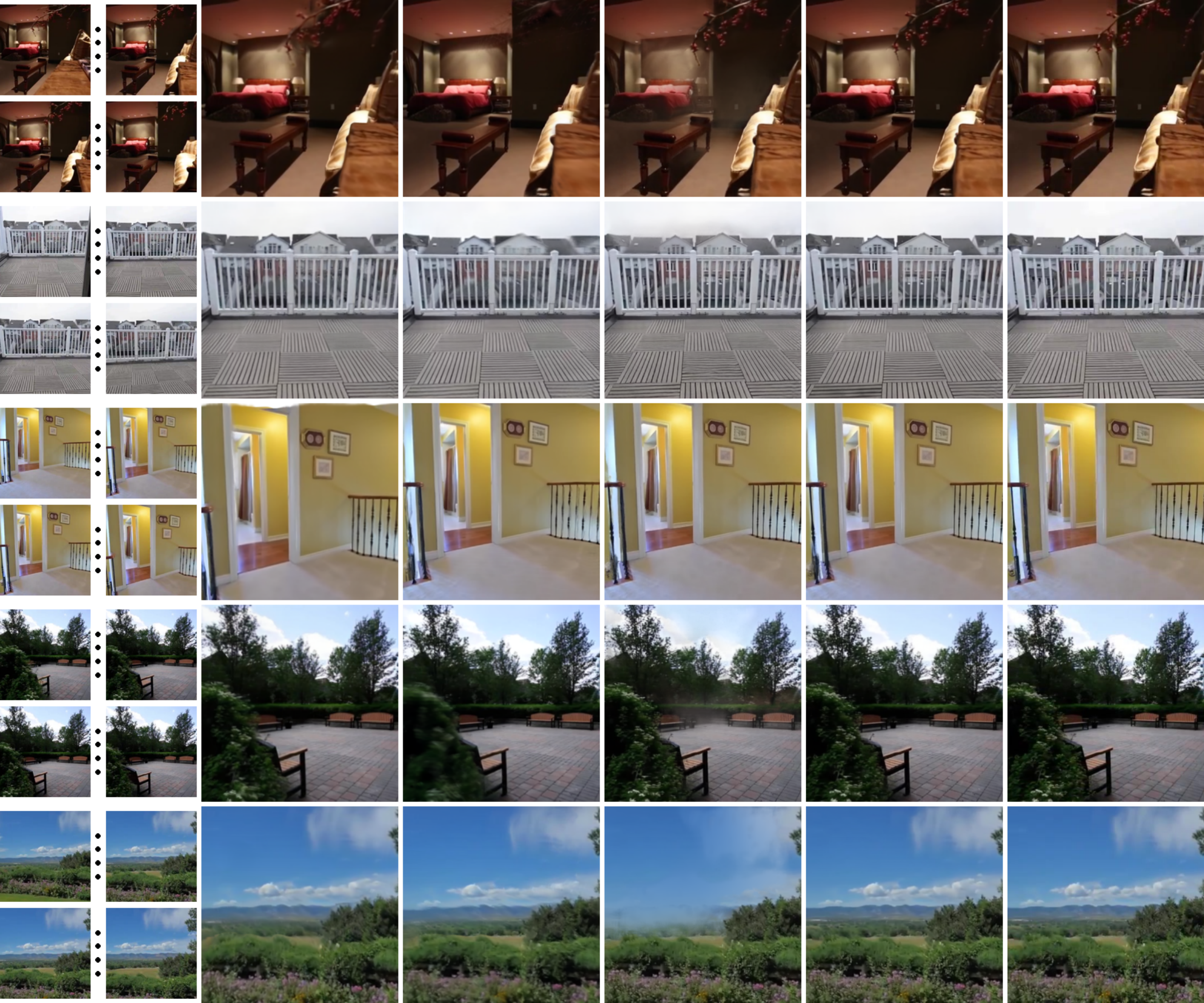}\\[1pt]
  \newcommand{\figcolf}[1]{\makebox[0.13078\textwidth][c]{\footnotesize #1}}%
  \newcommand{\figgapf}{\hspace{0.00306\textwidth}}%
  \figcolf{Inputs}\figgapf
  \figcolf{AnySplat}\figgapf
  \figcolf{WorldMirror}\figgapf
  \figcolf{VolSplat}\figgapf
  \figcolf{Ours}\figgapf
  \figcolf{Ground Truth}
  \vspace{-0.25cm}
  \caption{\textbf{More qualitative comparisons on RealEstate10K~\citep{zhou2018stereo}, 12 input views} (four shown). Checkpoints and evaluation follow \cref{fig:vis_re10k_12v}; selection is described in \cref{sec:appendix_vis_more}. Red boxes highlight local differences.}
  \label{fig:vis_re10k_more_12v}
\end{figure}

All qualitative figures use one selection procedure: per $64{\times}64$ window (stride 32) we compute each baseline's RMS gradient error minus ours, take each baseline's worst window, and rank frames by the weakest of these gaps, so every baseline must show a localized failure. Frames where any baseline beats \ours in PSNR or where \ours is below 26\,dB are discarded, and the top candidates are inspected zoomed. At 12 views (\cref{fig:vis_re10k_12v}) the 6-view failure modes persist: pose-free models wash out text, masonry, and branches, and \volsplat produces fog and floaters over distant or textureless regions. MVSplat, DepthSplat, and GGN have only 6-view checkpoints and are omitted.

\subsection{24-View Evaluation and Example Selection}
\label{sec:appendix_vis24v}

All methods run on the same 680 evaluation batches: \volsplat and \ours with their 24-view checkpoints, AnySplat ($448^2$, rendering at its predicted, scale-aligned target cameras) and WorldMirror ($518^2$, ground-truth camera priors) along their official NVS paths. Selection follows \cref{sec:appendix_vis12v} (3{,}352 of 5{,}440 frames pass the gate; top 30 inspected). Rows 2--4 are ranks 1, 3, and 4; rows 1 and 5 come from an earlier pass against \volsplat alone and also pass the three-baseline gate.

\subsection{Qualitative Results on ScanNet}
\label{sec:appendix_vis_scannet}

The ScanNet examples of \cref{fig:vis_scannet} were selected as in \cref{sec:appendix_vis12v}.

\subsection{More Qualitative Comparisons on RealEstate10K}
\label{sec:appendix_vis_more}

\cref{fig:vis_re10k_more_6v,fig:vis_re10k_more_12v,fig:vis_re10k_more_24v} use the methods, checkpoints, and protocol of \cref{fig:vis_re10k,fig:vis_re10k_12v,fig:vis_re10k_24v} and the selection of \cref{sec:appendix_vis12v} (the gate is passed by 2{,}474/5{,}592, 2{,}810/5{,}528, and 3{,}241/5{,}328 frames at 6/12/24 views), excluding scenes already shown and favouring distinct failure modes among the top 40: blurred or duplicated thin structures for pixel-aligned and pose-free baselines, streaks and fog for \volsplat.

\end{document}